\pdfoutput=1

\documentclass[11pt]{article}

\usepackage{acl}

\usepackage{times}
\usepackage{latexsym}

\usepackage[T1]{fontenc}

\usepackage[utf8]{inputenc}

\usepackage{microtype}

\usepackage{inconsolata}

\usepackage{amsmath}
\usepackage{booktabs}
\usepackage{enumitem}
\usepackage{xspace}
\usepackage{pifont}
\usepackage{graphicx}
\usepackage{subcaption}
\usepackage{diagbox}

\usepackage{wrapfig}

\usepackage{verbatim}
\usepackage{listings}
\usepackage{amsfonts}

\usepackage[linesnumbered,ruled,vlined]{algorithm2e}
\usepackage{multirow}
\usepackage{array}

\newcolumntype{C}[1]{>{\centering\arraybackslash}p{#1}}
\newcolumntype{L}[1]{>{\arraybackslash}p{#1}}

\newcommand{\meanstd}[2]{#1{\scriptsize $\pm$#2}}

\newcommand{\xmark}{\ding{55}}%

\newcommand{\name}{MLCopilot\xspace}

\title{\name: Unleashing the Power of Large Language Models\\in Solving Machine Learning Tasks}

\author{%
  Lei Zhang\thanks{Work done during Lei Zhang's internship at Microsoft Research Asia. Correspondence to Kan Ren.}$~~^1$ \quad Yuge Zhang$^1$ \quad Kan Ren$^2$ \quad Dongsheng Li$^1$ \quad Yuqing Yang$^1$ \\
  \textsuperscript{\rm 1}Microsoft Research, \textsuperscript{\rm 2}ShanghaiTech University \\
  \texttt{isleizhang@outlook.com, yugzhan@microsoft.com, renkan@shanghaitech.edu.cn}
}

\begin{document}

\maketitle

\begin{abstract}
The field of machine learning (ML) has gained widespread adoption, leading to significant demand for adapting ML to specific scenarios, which is yet expensive and non-trivial. The predominant approaches towards the automation of solving ML tasks (\emph{e.g.,} AutoML) are often time-consuming and hard to understand for human developers. In contrast, though human engineers have the incredible ability to understand tasks and reason about solutions, their experience and knowledge are often sparse and difficult to utilize by quantitative approaches. In this paper, we aim to bridge the gap between machine intelligence and human knowledge by introducing a novel framework \name\footnote{Examples and code available at \url{https://github.com/microsoft/CoML}}, which leverages the state-of-the-art large language models to develop ML solutions for novel tasks. We showcase the possibility of extending the capability of LLMs to comprehend structured inputs and perform thorough reasoning for solving novel ML tasks. And we find that, after some dedicated design, the LLM can \emph{(i)} observe from the existing experiences of ML tasks and \emph{(ii)} reason effectively to deliver promising results for new tasks. The solution generated can be used directly to achieve high levels of competitiveness. 

\end{abstract}

\section{Introduction}
Past decades have witnessed a great advance and rapid development of machine learning (ML), but ML algorithms are still notoriously hard to configure \cite{hutter2019automated}.
For specific tasks, configuring and conducting corresponding ML solutions is non-trivial, which thus requires extensive human labor.
Many challenges arise in developing practical ML solutions.
First, it is of large human efforts considering the large space of ML solutions, such as feature engineering, model design, optimization details, etc. 
Second, ML algorithms are sensitive to even minor changes of the task context.
As a result, even the same algorithm may need to be reconfigured for different application tasks.
Last but not least, transferring successful experiences across different tasks is also intractable, which demands  high-level reasoning abilities of human experts to derive reasonable solutions for novel tasks.

The predominant approaches that relieve the human effort of algorithm configuration have been some automation mechanisms such as AutoML (Automated Machine Learning) \cite{hutter2019automated}.
One major branch of AutoML formulates the problem as black-box optimization, and resorts to some optimization approaches such as Bayesian optimization (BO) \cite{frazier2018tutorial} to solve it.
Though the obtained results are shown to be promising, it is time-consuming to spawn multiple trials, especially for large datasets and complex tasks.
Moreover, AutoML does not follow the natural pattern of ML development that humans are accustomed to, which leaves a huge gap for humans to understand and control the whole process.
Specifically, it is either difficult to explain the behavior of auto-tuning, or intractable to incorporate human prior such as the knowledge of the model architectures into the process, making it less flexible for human developers.
Furthermore, the ML solutions derived by these optimization-based methods may \textit{only} fit to the specific domains, and the transferring ability of these results also remains an open problem \cite{chen2022towards,yan2022privacy}.

Contrarily, we notice two tendencies in how humans approach an ML task. Instead of jumping into solving the new task directly, humans often try to comprehend the task at hand and draw from their past experiences on relevant tasks.
Additionally, humans recall their knowledge, which may have came from a textbook or prior experiences. This process differs significantly from the automated approach mentioned earlier, which leads us to a natural question: 
\textit{can we leverage both machine intelligence and human design patterns to improve our ability to solve ML tasks?}
The advances of Large Language Models (LLM) \cite{brown2020language,chowdhery2022palm,ouyang2022training} have illustrated tremendous promising performance in mimicking human behaviors on conversation-based tasks.
It seems plausible to utilize the power of LLM to address ML problems in a more human-like way. 

Nevertheless, several challenges remain when incorporating LLMs to achieve this goal.
First, we discovered that LLMs have trouble performing ML tasks based solely on the task description, in which case the performance is no better than random generation.
Attempting to leverage historical ML experience, we found that the data often reside in heterogeneous formats (\emph{e.g.,} code, configs and logs), which need to be canonicalized into formats that are acceptable to LLMs.
Moreover, the amount of information that can be incorporated into in-context learning~\cite{brown2020language} is quite limited, and thus some retrieval strategy is desired to make the best out of it.
Finally, deriving a ML solution based on historical experience is in its essence a mathematical thinking and logical reasoning problem~\cite{patel2021nlp}, which necessitates some mechanisms to reasoning over knowledge.

In this paper, we explore and present a novel framework \textit{\name}, which leverages LLMs to suggest solutions for novel real-world ML tasks, based on the existing experiences from historical tasks.
We decompose the problem into offline and online stages.
In the offline stage, \name \emph{canonicalizes} historical data and creates an experience pool. LLMs are then used to \emph{elicit} valuable knowledge from historical experience.
In the online stage, \name retrieves experiences from the most relevant tasks from the experience pool, given the description of the target task. It then interacts with LLMs to obtain multiple suggested ML solutions in one round.
We demonstrate that, with a well-designed framework, LLMs can not only elicit meaningful knowledge from historical experiences but also provide reasonable and competitive ML solutions for novel tasks.

Our work presents a three-fold contribution, which can be summarized as follows.
    \emph{(i)} To the best of our knowledge, we are the first to utilize LLMs as a tool to generate solutions for new ML tasks. 
    \emph{(ii)} A novel retrieve-and-prompt framework has been proposed to solve ML tasks almost instantaneously, \textit{without} any time-consuming searching or optimization.
    \emph{(iii)} We leverage the text understanding and generation capabilities of LLMs to produce \emph{interpretable}\footnote{The concept of ``interpretability'' in the context of our work primarily pertains to the human-readable knowledge generated by LLMs, which serves as a transparent reference for decision-making.} results for ML tasks. This approach has shown comparable or even better performance on a variety of real-world ML benchmarks.

\section{Related Work}

\begin{table*}[t]
    \centering

    \resizebox{\linewidth}{!}{%
    \begin{tabular}{l|l|l}
    \hline
    \textbf{Term} & \textbf{Definition} & \textbf{Example} \\
    \hline
    Task $T$ & ML problem to solve (optionally with constraints). & Find an optimizer for a ResNet on ImageNet dataset. \\ \hline
    Solution space $\mathbb{S}$ & Solution hypothesis space to the task. & Optimizer: \{Adam, SGD\}; Learning rate: $[10^{-6}, 0.1]$. \\ \hline
    Solution $S$ & One particular choice within solution space. & 2-layer ResNet with SGD optimizer using LR $10^{-3}$. \\ \hline
    Experience $E$ & Successful solutions on historical tasks. & \begin{tabular}{@{}l@{}} SGD with lr 0.024 achieves 76.2\% accuracy for \\ ResNet on ImageNet. \end{tabular} \\ \hline
    Knowledge $K$ & High-level information acquired from experiences. & \begin{tabular}{@{}l@{}} Usage of small LR makes training slower but could \\ yield better final result. \end{tabular} \\
    \hline
    \end{tabular}%
    }
    \vspace{-0.1in}
    \caption{
    Terminologies used throughout this paper.
    }
    \label{tab:terminologies}

    \vspace{-0.1in}
\end{table*}

\subsection{Large Language Models}
Large language models (LLMs) are neural networks of significant sizes (typically containing tens or hundreds of billions of parameters). They have gained the incredible ability of processing and generating natural languages, due to the training on massive amounts of text data~\cite{Radford2018GPT,Radford2019GPT2,brown2020language}. 
Studies show that LLMs beyond a certain scale have ``emergent abilities''~\cite{wei2022emergent}, and perform remarkably well in applications such as chatbots, machine translation, and text summarization \cite{zhao2023survey,touvron2023llama}.

While LLMs have illustrated superior performance on natural language understanding and human-like text generation, they are still quite limited for complicated tasks that require reasoning~\cite{huang2022towards} and mathematical skills~\cite{patel2021nlp,thawani2021representing,han2022luna,saxton2019analysing}.
The stream of task automation~\cite{lu2023chameleon,shen2023hugginggpt} investigated a general approach to decompose a task into a sequence of sub-tasks, but they are orthogonal to our work, since they did not take into past experience or knowledge from other tasks when planning a new task.

\subsection{Machine Learning and AutoML}
Machine learning (ML) is a subfield of artificial intelligence (AI) that involves developing optimization algorithms that can learn from data and make predictions \cite{bishop2006pattern} or decisions \cite{sutton2018reinforcement}.  Although ML has been successful in many real-world applications, designing an effective ML solution for a new task can be challenging due to the numerous design choices required. AutoML~\cite{hutter2019automated} emerges as an approach to alleviate the manual effort involved. Popular methodologies include neural architecture search (NAS)~\cite{pham2018efficient}, meta-learning~\cite{andrychowicz2016learning}, and Bayesian optimization~\cite{frazier2018tutorial}.

AutoML is able to reach beyond-human levels in solving ML tasks, but it still faces a few drawbacks. First, most AutoML methods require many rounds of trial-and-error, which can be time-consuming. Second, AutoML typically searches from scratch for a new task and neglects the experience on previous tasks. Finally, most AutoML methods are not interpretable due to their black-box nature, which excludes human understanding. Some methods may address one or two of the drawbacks, but not all of them.  For example, the stream of transferrable AutoML research seeks to leverage past experience to assist in searching for new tasks~\cite{bardenet2013collaborative,wistuba2016two,mittal2020hyperstar,yan2022privacy,wang2021flaml}, but they lack interpretability and most of them only work for specific types of tasks. A recent study~\cite{chen2022towards} aims to use Transformer model~\cite{vaswani2017attention} with large-scale pretraining to deal with broader types of tasks, but it is still non-interpretable and a cost search remains required for new tasks. Most recently, \cite{zheng2023gpt4} tried to search for neural architectures using GPT-4~\cite{openai2023gpt4}. It prompts LLM to explain rationales, but it only explores model architectures, and still requires costly trial-and-error.

\section{Preliminaries}

The goal of \name is to assist humans in solving complex ML problems. 
Generally speaking, given a \textit{task} which is a real-world problem for ML models to tackle, the goal of ML development is to conduct a concrete \textit{solution}. The solution can be either a pipeline, configuration, or code snippet, based upon which a concrete ML model could be learned to handle the target task.
The solution is also a particular sample within a complicated \textit{solution space} that involves various design choices.
These choices are mutually correlated and the outcome of different alternatives often influences the others and eventually affects the final performance of the overall ML solution.

To create reasonable ML solutions for new tasks, we can draw on \textit{experiences} from previous relevant tasks. \name is designed to use historical experiences for \textit{knowledge} elicitation and effectively conduct effective solutions for the given novel ML task (details in \S~\ref{sec:methodology}). To improve comprehension and clarity, we summarize the terminologies with descriptions and examples in \autoref{tab:terminologies}.

\section{\name}
\label{sec:methodology}

In this section, we present \name, with the formulation of the main problem and the overall architecture of our method.
Then, we will describe some key components of \name in detail, including target task description, retrieval, canonicalization, and knowledge elicitation.

\begin{figure*}
\centering
\begin{minipage}{.65\textwidth}
\centering
\includegraphics[width=\linewidth]{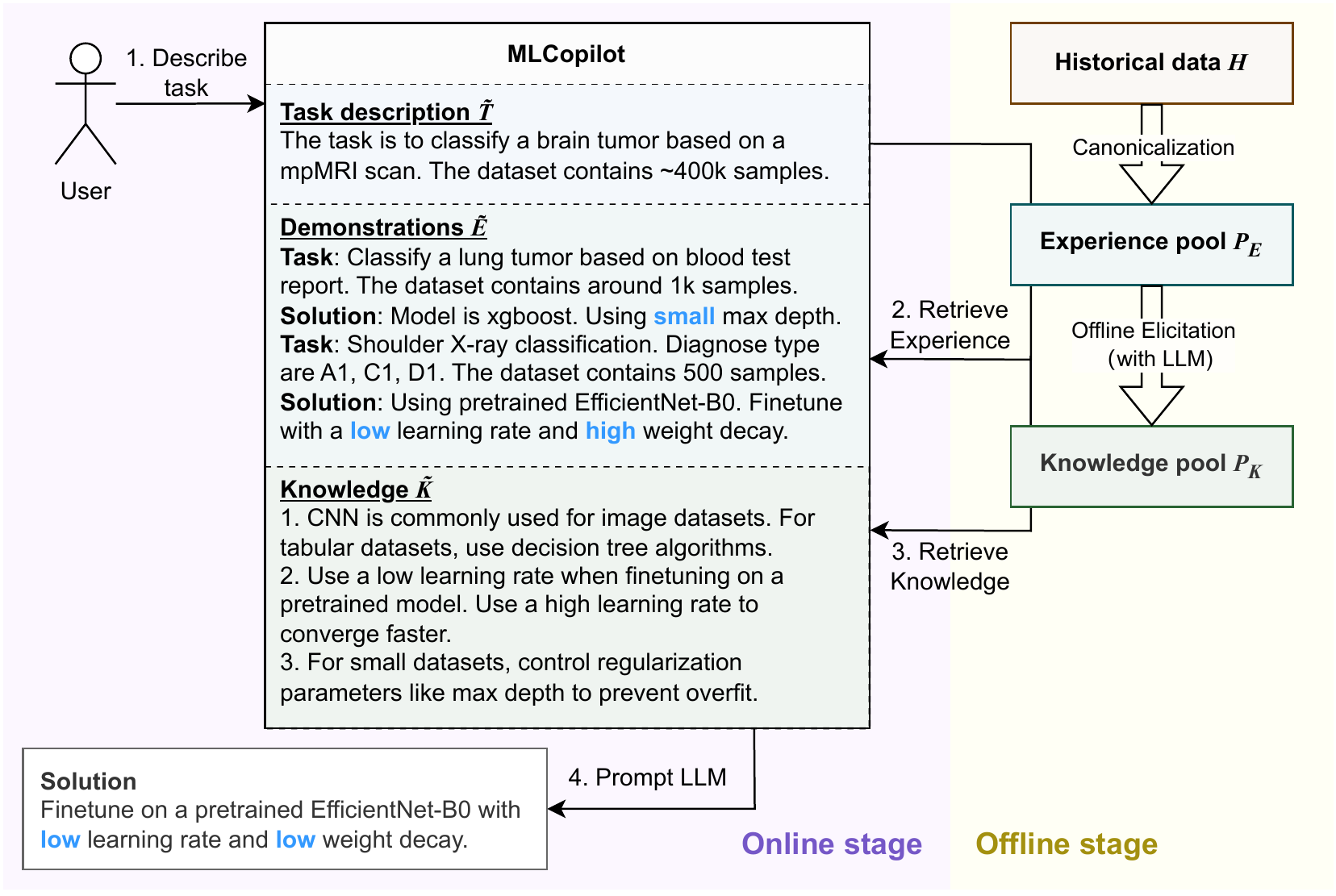}
\caption{Overview of \name. MLCopilot has offline and online stages. During the offline stage, it creates pools of experience and knowledge. In the online stage, it retrieves experience and knowledge based on the novel task description. Finally, MLCopilot invokes LLM and returns solutions.}
\label{fig:mlcopilot1}
\end{minipage}%
\hspace{0.03\textwidth}%
\begin{minipage}{.285\textwidth}
\centering
\includegraphics[width=.95\linewidth]{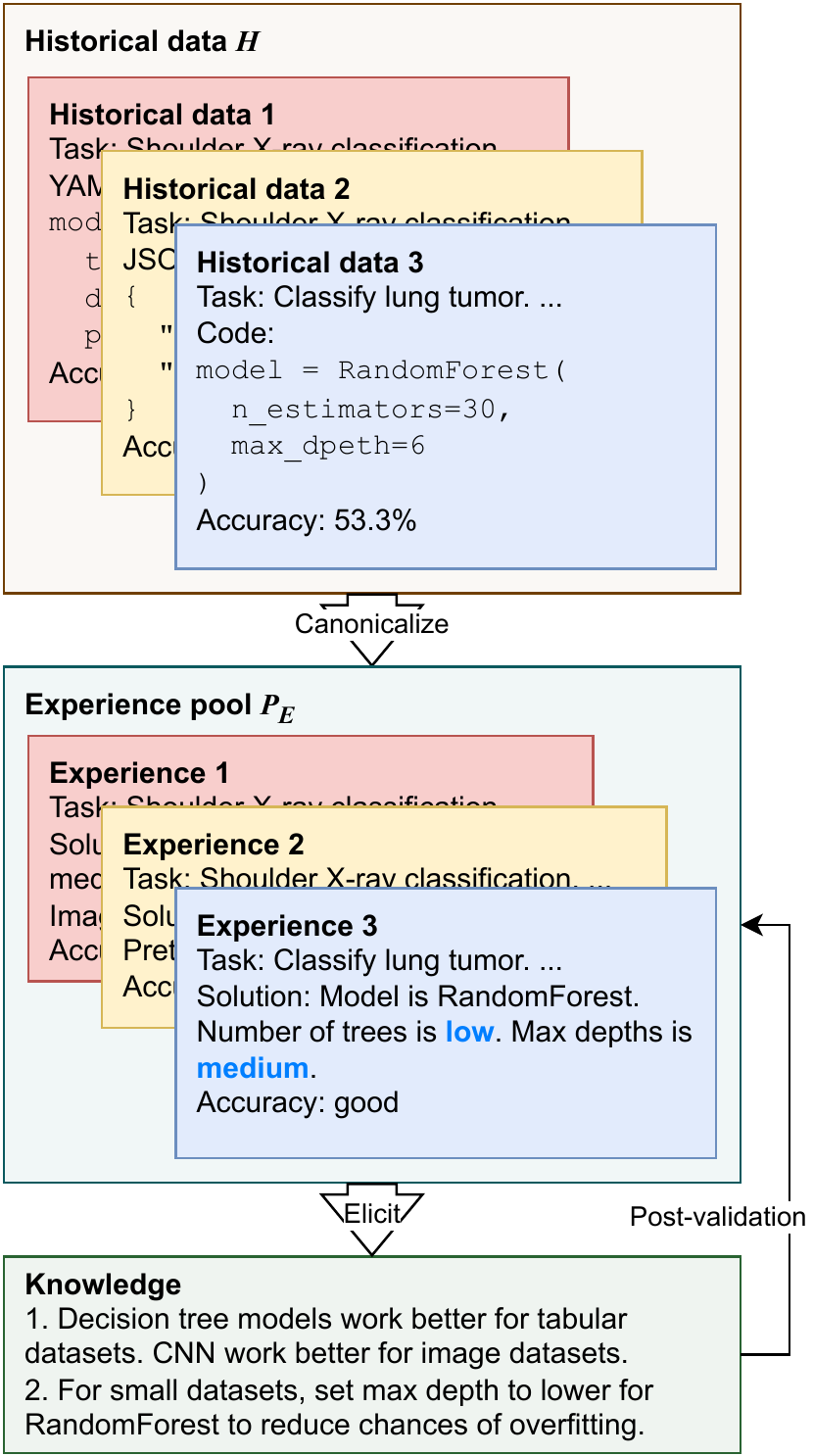}
\caption{Offline stage: canonicalization, knowledge elicitation.}
\label{fig:mlcopilot2}
\end{minipage}
\vspace{-.1in}
\end{figure*}

\subsection{Overall Framework}

As discussed previously, to unleash the power of LLMs in solving complex ML tasks, explicitly leveraging historical experience is crucial.
However, utilizing past experience is not straightforward considering the heterogeneous data format and the huge number of records.
Therefore, our technical design mainly focuses on addressing two problems:
\emph{(i)} how to comprehend and exploit the abundant raw experiences;
\emph{(ii)} how to effectively solve ML tasks based on the result of (i).

The main idea behind \name
is \textit{knowledge-based reasoning}, that is to leverage LLMs to conduct reasoning and task solving based on the previous knowledge, which has been analyzed and elicited from past experiences.
To this end,
\name contains two stages, including offline and online parts, both of which have been visually illustrated in \autoref{fig:mlcopilot1}.
In the offline stage, LLM has been incorporated to analyze the canonicalized historical experience data and elicit useful knowledge.
And in the online stage, the user will query \name, which is also built upon LLM, to obtain a suitable ML solution for the novel task.

\subsubsection{Offline Stage: Understanding and Reasoning}

We first present the data settings and describe the corresponding preprocessing procedure briefly.
Let $H = \{D_1, \ldots, D_{N_H}\}$ be the raw historical data with $N_H$ previous records. 
The $i$-th record $D_i$ is defined as a three-element tuple $\langle T_i, S_i, M_i \rangle$ which contains a task $T_i \in \mathbb{T}$,
a solution $S_i \in \mathbb{S}$, and the evaluated metric performance $M_i$, \emph{e.g.,} classification accuracy.

Note that, the historical data $H$ may have heterogeneous and diverse formats. 
For example, the task can be described in the natural text, while the solution can be a JSON configuration, a row of tabular data, or code snippets. 
An experience pool $P_E$ is constructed, to canonicalize the data and store them as experiences $P_E = \{E_1,\ldots,E_{N_E}\}$, where $E_j =  \mathcal{C}(D_j) $, and $\mathcal{C}(\cdot)$ is the canonicalization function.
For the simplicity of notations, we assume all the solutions within $P_E$ come from a universal solution space. 
It is easy to extend the framework to scenarios with multiple solution spaces.

Knowledge, is high-level information acquired from the experience~\cite{dictionary1989oxford}, 
and we leverage LLM to elicit knowledge from the constructed experience pool, in the offline stage. 
Knowledge is the easy-to-understand summarization of previous ML experiences, which will further be utilized when inferring the final solution in the online stage.
To be specific, a subset of experience is first sampled from the experience pool, 
then a LLM is incorporated to read and understand the experience data, allowing us to ``elicit'' knowledge $K$ from it.
The process of elicitation is formulated as $K = \mathcal{I}_K(P_E;\textsc{Llm})$, which is an iterative process by interacting with LLM along with post-validation on the obtained knowledge.
The detailed process of elicitation is discussed in \S~\ref{sec:elicitation}.
All the generated knowledge is stored in a knowledge pool $P_K = \{K_1, \ldots, K_{N_K}\}$ with totally $N_K$ items.

The obtained experience pool and knowledge pool will be further utilized by \name in the online stage, to conduct reasonable, promising, and competitive ML solutions for novel tasks.

\subsubsection{Online Stage: Retrieving and Solving}

The online stage of \name aims to conduct reasoning and task solving based on the off-the-shelf information obtained from the offline stage. 
Specifically,
given the user query with a task description, \name will respond with the corresponding reasonable ML solutions via retrieving relevant experiences and knowledge, and interacting with LLM by a curated prompt, in one round.

When a user comes with a novel target task $\tilde{T}$, which has never been seen in history, \name first retrieves the relevant experiences of other relevant tasks as demonstrations $\tilde{E} = \mathcal{R}_E(\tilde{T}, P_E)$, where $\mathcal{R}_E(\cdot)$ is the retrieval functions for the experience pool. 
It also retrieves knowledge $\tilde{K} = \mathcal{R}_K(\tilde{T}, P_K)$ to guide the response on the new task, where $\mathcal{R}_K(\cdot)$ is the retrieval functions for the knowledge pool. 
\name finally generates a recommended solution by invoking LLM once: $\tilde{S} = \textsc{Llm}(\tilde{T}, \tilde{E}, \tilde{K})$.

The framework is shown in \autoref{fig:mlcopilot1}, where we illustrate an example of how \name handles a task to classify a brain tumor by leveraging previous experiences and knowledge. 
Next we will introduce the dedicated components in detail.

\subsection{Task Description in Natural Language}

Firstly, we show how the target task is described in our framework, which is the input to \name. 
The prior works~\cite{feurer2015efficient,wang2021flaml} usually use meta-features designed by humans for specific types of tasks (\emph{e.g.,} the number of samples in the dataset) to describe a task, so as to ease the difficulty of comprehending the task. 
However, such design might degenerate the ability of LLM to generalize to new types of tasks.
We believe that
task description in natural language is more straightforward to users.
It is also agnostic to task types, and does not require heuristics to design meta-features.
As such, we adapt the task description without any feature engineering, and users can freely describe dataset names, characteristics, domain-specific constraints, and more.
Furthermore, our experiments (\S~\ref{sec:robustness}) illustrated that incorporating a natural language user interface helps recall and leverage previous knowledge contained in the training corpus of LLMs.

\subsection{Retrieval}

The retrieval technique has been used to \emph{(i)} gather some demonstrations of the historical ML solutions to the relevant tasks and \emph{(ii)} apply useful knowledge previously to further motivate and prompt the LLM to better solve the target ML task.

We first discuss how to retrieve experience $\mathcal{R}_E$ as demonstrations $\tilde{E}$.
Intuitively, the most helpful experience in solving a new task should come from the most relevant tasks. 
The key question then becomes how relevance is defined. 
To this end, we first embed the task description by invoking a language model $\mathcal{E}$ (\emph{e.g.,} GPT-3 \cite{brown2020language}) to generate an
embedding vector of the textual content. 
Given a new task $\tilde{T}$, \name retrieves the most relevant historical tasks from the experience pool $P_E$ by calculating the cosine similarity between the embeddings of the new task and the stored tasks.
The corresponding experience to these embeddings will serve as demonstrations $\tilde{E}$, as calculated as

\vspace{-.3in}
$$
\tilde{E} = \mathcal{R}_E(\tilde{T}, P_E) = \underset{\langle T, S, M \rangle \in P_E}{\arg \textrm{top-}k} \left( \frac{\mathcal{E}(T) \cdot \mathcal{E}(\tilde{T})}{ |\mathcal{E}(T)| \cdot |\mathcal{E}(\tilde{T})| }  \right) ~,
$$
where the most relevant $k$ entries of experience will be retrieved for subsequent demonstration.

The retrieval of knowledge $\mathcal{R}_K$ is based on matching of solution space -- retrieving \textit{all} the knowledge that are elicited from the same solution space as the new task, from the knowledge pool $P_K$.
This simplicity of $\mathcal{R}_K$ is due to the fact that the knowledge produced in the offline stage of \name is concise and of high quality, whose generation procedure will be discussed in \S~\ref{sec:elicitation}.

\subsection{Canonicalization}
\label{sec:Canonicalization}

As mentioned previously, the data of raw ML experience are heterogeneous and of diverse formats.
While some of them (\emph{e.g.,} task descriptions) have already been in natural text format, the ML solutions and the corresponding metric performance are often expressed in structured configurations, tabular formats, or programming languages. 
More importantly, they might even contain a lot of numbers, which language models or even LLMs are not good at processing and understanding~\cite{thawani2021representing,han2022luna,saxton2019analysing}.
To better unleash the power of LLM, we canonicalize all the data to express it in natural language.

The essential part of canonicalization is to convert the raw data into a well-formed natural language, as shown in the left part of \autoref{fig:mlcopilot2}. 
Other than unifying the solutions in diverse formats, a crucial technique is number discretization which avoids feeding numbers to LLM directly. 
We follow \cite{thawani2021representing} that discretizes continuous numerical data into several intervals, and mapping each value within each interval to the same discrete value.
To minimize performance loss, we discretize each numerical value based on the corresponding distribution and percentile points.
More details can be found in \S~\ref{sec:implementation-details}.

\subsection{Knowledge Elicitation}\label{sec:elicitation}

With the canonicalized experience stored in pool $P_E$, \name can then elicit knowledge, to better support the online stage for solving novel ML tasks. 
It is important to note that knowledge elicitation occurs offline, prior to serving user tasks.
The approach involves the following steps: \emph{(i)} constructing a prompt that consists of a \emph{random subset} of the experience pool (to avoid bias towards certain tasks),
along with an inquiry that asks for analysis and summary;
\emph{(ii)} sending the prompt to LLMs to generate a knowledge ``candidate'';
\emph{(iii)} validating the candidate on experience pool. 
The flow of this process is illustrated in the right part of \autoref{fig:mlcopilot2} (pseudo-code in \S~\ref{sec:algorithms}).

We elaborate on the validation step, which we call automated \emph{post-validation} after requesting knowledge from the LLM. 
This step is designed to alleviate the hallucination issue~\cite{ji2023survey} and raise the quality of generated knowledge.
It tests the knowledge by using it to solve a set of \emph{validation tasks}. 
If the generated knowledge is found invalid (\emph{e.g.,} due to hallucination), it adjusts the generation settings, such as the order of experiences in the prompt, tone of hypophora questions, and parameters of LLM invocation, and let LLM regenerate knowledge. 
This iterative process continues until the performance on the validation tasks has been converged,
or the invocation has reached the maximum number. 
This process can be represented formally as $K = \mathcal{I}_K(P_E;\textsc{Llm})$.

We argue that our knowledge elicitation is novel and different from prior works of knowledge extraction~\cite{zhang2022deepke} or knowledge generation~\cite{yu2022generate,lu2023chameleon} in natural language processing.
Firstly, our knowledge is obtained from heterogeneous resources using general text completion models, without requiring predefined templates or complicated pipelines.
Secondly, acquiring knowledge for ML tasks requires analysis, summarization, and high-level reasoning, which is significantly more challenging than simply extracting simple facts~\cite{zhang2022deepke}.
Finally, the knowledge is anchored in experience data, rather than purely based on LLM's pre-training corpus~\cite{yu2022generate}, which makes the framework scalable to new scenarios.

Knowledge elicited by \name can be beneficial not only for LLMs but also for human ML practitioners. Since the knowledge is expressed in natural language, it could potentially serve as a cookbook for ML developers. In an effort to share our findings and inspire future ML research, we have released all the knowledge obtained so far (see \S~\ref{sec:appendix-knowledge}). Hopefully this will reveal some of the ``secret sauce'' behind how ML works and promote knowledge sharing within the community.

\section{Experiment}

We evaluate \name on a series of benchmarks, aiming to answer the following research questions:
\emph{(i)} Can \name outperform traditional approaches or simple interactions with LLMs?
\emph{(ii)} How important are individual techniques in \name, \emph{e.g.,} knowledge and experience?
\emph{(iii)} Is the elicited knowledge informative and reasonable?

\subsection{Experiment Setup}
\label{sec:implementation-details}

\textbf{Implementation details.} The current implementation of \name involves maintaining dedicated experience and knowledge pools for each solution space. The historical data is sourced from the benchmarks described below, while task descriptions are crawled from benchmark websites. Numerical values in the data are discretized into five levels: ``very low'', ``low'', ``medium'', ``high'', and ``very high''. The precise value of each level is determined by analyzing the statistics of the best solutions within the solution space (see detailed analysis in \S~\ref{sec:ablation}). We interact with the general-purpose GPT-3.5 model\footnote{\url{https://platform.openai.com/docs/models/gpt-3-5}} (code-named ``text-davinci-003'', \textit{without} additional fine-tuning), and ``text-embedding-ada-002'' to obtain embeddings for task descriptions. (Results with other LLMs can be found in \S~\ref{sec:choice-of-llms}.)
The temperature is set to 0 to minimize randomness. Additional details regarding prompt design can be found in \S~\ref{sec:appendix-prompt-design}.

\noindent \textbf{Benchmarks.}
We selected benchmarks that have established a predetermined solution space for all possible solutions and provided performance metrics for all the solutions in the solution space (either through a lookup table or surrogate).
We conducted experiments using \name on three ML benchmarks: HPO-B~\cite{arango2hpo}, PD1~\cite{wang2021pre}, and HyperFD~\cite{yan2022privacy}. These benchmarks comprise numerous ML tasks and datasets, covering a broad spectrum of scenarios such as tabular data classification and regression, image classification, and object detection. 
Details can be found in \S~\ref{sec:appendix-benchmark}.

\noindent \textbf{Evaluation metrics.} In each experiment, every compared method makes three attempts to predict successful solutions for an unseen task. Solutions are evaluated in the order they were suggested. Metric@$t$, where $t \ge 1$, is defined as the best metric performance achieved among the first $t$ suggested solutions. The reported performance metrics are averaged over at least 5 random seeds.

\begin{table*}[t]
\resizebox{\linewidth}{!}{%
\begin{tabular}{c|ccc|ccc|ccccccc}
\hline
\multirow{2}*{\textbf{Method}} & \multicolumn{3}{c|}{\textbf{HPO-B $\uparrow$}} & \multicolumn{3}{c|}{\textbf{PD1 $\uparrow$}} & \multicolumn{6}{c}{\textbf{HyperFD (Rank $\downarrow$ ~ AP $\uparrow$)}} \\
\cline{2-13}
& \textbf{nAcc@1} & \textbf{nAcc@2} & \textbf{nAcc@3} & \textbf{nAcc@1} & \textbf{nAcc@2} & \textbf{nAcc@3} & \textbf{Rank@1} & \textbf{Rank@2} & \textbf{Rank@3} & \textbf{AP@1} & \textbf{AP@2} & \textbf{AP@3} \\
\hline
Random & 54.70 & 60.70 & 64.80   & -0.86 & -0.08 & 0.39 & 109.55 & 73.16 & 54.79 & 90.76 & 91.20 & 91.38 \\
Constant & 72.85 & 74.61 & 75.02  & 1.27  &  1.56 & 1.59 & 78.00 & 54.33 & 49.25 & 91.13 & 91.23 & 91.28 \\
TST-M & 72.73 & 74.44 & 74.56 & 1.10 & 1.35 & 1.40 & 57.67 & 43.50 & 42.67 & 91.22 & 91.37 & 91.38 \\ 
HyperSTAR & 67.37 & 68.14 & 68.71 & 1.10 & 1.27 & 1.34 & 97.75 & 72.97 & 52.03 & 90.78 & 91.05 & 91.25 \\
ASKL & 77.01 & 81.76 & 85.02 & 1.26 & 1.29 & 1.44 & 92.58 & 64.67 & 51.25 & 90.93 & 91.15 & 91.34 \\
FLAML & 77.84 & 82.95 & 88.06 & 1.28 & 1.31 & 1.58 & 66.42 & 43.33 & 31.83 & 91.09 & 91.29 & 91.33 \\ 
HyperFD & -- & -- & -- & -- & -- & -- & 56.97 & 47.91 & 31.75 & 91.17 & 91.26 & 91.44 \\
\hline
 LLM-ZS & 61.37 & 79.41 & 80.56 & -1.03 & 1.25 & 1.26 & 119.25 & 90.42 & 41.00 & 90.69 & 90.96 & 91.38 \\
 LLM-FS & 78.93 & 83.10 & 89.73 & 0.43 & 0.57 & 0.62 & 66.69 & 52.43 & 40.98 & 91.26 & 91.40 & 91.48 \\
\rowcolor{gray!30} \name & 81.59 & 83.23 & 90.72 & 1.48 & 1.54 & 1.62 & 59.74 & 38.67 & 25.58 & 91.38 & 91.60 & 91.66 \\
\hline
\end{tabular}
}

\caption{Main results on HPO-B, PD1 and HyperFD. \textbf{nAcc, AP:} the higher the better. \textbf{Rank:} the lower the better.}
\label{tab:performance-all}

\end{table*}

\subsection{Main Results}

We show the performance of \name in \autoref{tab:performance-all}.
Baselines we compared with include:

\begin{itemize}[leftmargin=1em]
    \item Traditional AutoML or meta learning methods, including Random, ASKL~\cite{feurer2015efficient}, Constant~\cite{bardenet2013collaborative, kotthoff2019auto}, TST-M~\cite{wistuba2016two}, HyperSTAR~\cite{mittal2020hyperstar}, HyperFD~\cite{yan2022privacy} and FLAML-Zero~\cite{wang2021flaml}. Details described in \S~\ref{sec:appedix-baseline}.
    \item \textbf{LLM-ZS} directly prompts LLM to generate a zero-shot solution based solely on the task description, which is similar to using tools such as GitHub Copilot\footnote{\url{https://github.com/features/copilot}} or Amazon CodeWhisperer\footnote{\url{https://aws.amazon.com/codewhisperer/}}.
    \item \textbf{LLM-FS} uses the few-shot prompt technique~\cite{brown2020language} by adding some demonstrations to the prompt to enable in-context learning. The demonstrations are randomly selected from our canonicalized experience pool. Unlike \name, LLM-FS does not have access to advanced techniques such as experience and knowledge retrieval.
\end{itemize}

\begin{table*}[t]
    \small
    \centering

    \resizebox{\linewidth}{!}{%
    \begin{tabular}{ccccccc}
    \toprule
    
    \multirow{2}{*}{\diagbox{\textbf{Retrieved by}}{\textbf{Pipeline}}} &  \multicolumn{2}{c}{\textbf{HPO-B} $\uparrow$} & \multicolumn{2}{c}{\textbf{PD1} $\uparrow$} &\multicolumn{2}{c}{\textbf{HyperFD} $\downarrow$}  \\ \cline{2-3} \cline{4-5} \cline{6-7}
    & ASKL & \name & ASKL & \name & ASKL & \name \\ \midrule 
    Text embedding &\meanstd{75.34}{0.00} & \meanstd{81.59}{0.94}  & \meanstd{1.40}{0.00} & \meanstd{1.48}{0.06} & \meanstd{79.17}{0.00} & \meanstd{59.74}{1.89} \\ 
    Meta-feature  & \meanstd{80.61}{0.00} & \meanstd{83.29}{1.46} & \meanstd{1.02}{0.00} & \meanstd{1.41}{0.09} & \meanstd{107.67}{0.00} & \meanstd{50.49}{6.38} \\ 
    Random  &\meanstd{73.67}{2.51} & \meanstd{78.00}{2.82} & \meanstd{0.06}{0.25} & \meanstd{1.37}{0.12} & \meanstd{84.23}{14.84} & \meanstd{57.95}{10.19} \\ 
     \bottomrule
    \end{tabular}%
    }
    \vspace{-0.1in}

    \caption{Comparison of approaches to retrieve experience (\emph{i.e.,} based on what measures to retrieve the experience) and to consume the retrieved experience (ASKL: directly use the solutions for retrieved tasks on the new task; \name: use the retrieved experience as demonstrations along with knowledge to prompt LLM).}
    \label{tab:ablation-retrieval}

    \vspace{-0.1in}
    \end{table*}

\name achieved the highest normalized accuracy (nAcc) across all three trials. The improvement is particularly significant for the first attempt (nAcc@1). It is remarkable that LLM-FS has already surpassed all the traditional baselines, suggesting the large capability of LLMs on ML tasks.

On PD1, Normalized accuracy (nAcc) are in range $[-2,2]$ following the setting of \cite{wang2021pre}. \name remains the best out of all methods compared. Notably, ``Constant'' baseline almost outcompetes all other baselines, which casts doubt on the effectiveness of the task similarities measured by other baselines. Meanwhile, both LLM-ZS and LLM-FS fail on PD1, indicating PD1 is more challenging for LLMs.

For HyperFD, Following \cite{yan2022privacy}, we use average precisions (AP) (the higher the better) and rankings (within $[1, 216]$, the lower are better) to measure the performance.
Similar to what was observed in HPO-B, LLM-FS achieves comparable performance to most baselines with a few demonstrations. It is expected that the performance of LLM-FS would improve with the inclusion of techniques from \name. However, it is worth noting that HyperFD is a private benchmark and its benchmark was released after the knowledge cutoff of GPT-3.5, making it unlikely that LLM has memorized the best solutions on this benchmark.

\subsection{Ablation study}\label{sec:ablation}

In the ablation study, we use nAcc@1 (or Rank@1) as the main metric for comparisons.

\noindent \textbf{Study of retrieval. }
The first question we are trying to answer is whether retrieving experience and knowledge are necessary -- what happens if either of them is missing from the prompt sent to LLM? The results are shown in \autoref{tab:ablation-experience}. While the absence of knowledge leads to a reduction in performance, the absence of demonstrations leads to a complete collapse.
Examining the knowledge generated (\S~\ref{sec:appendix-knowledge}), we found it often contains vague claims such as ``\emph{The size of the dataset can influence the configuration of eta}'' (HPO-B, Space 5971). The knowledge did not clarify what is the ``\emph{influence}'', which is why experience is still much needed even with the presence of knowledge.

We then compare different retrieval methods (``Pipeline \name'' columns in \autoref{tab:ablation-retrieval}).
\name retrieves the most relevant tasks based on the \emph{embedding of textual description}. Alternatively, we can \emph{(i)} measure similarities based on \emph{meta-features}; \emph{(ii)} simply retrieve experiences \emph{randomly}. Shown in  \autoref{tab:ablation-retrieval}, both meta-feature and text embedding consistently outperform random retrieval.

When choosing between meta-features or text embedding, we believe that the latter has demonstrated advantages over manually designed meta-features. This is partly due to the fact that the performance of meta-features depends largely on the quality of their design. While meta-features have been shown to be promising for tabular datasets where they are well-studied and carefully designed~\cite{feurer2015efficient,wang2021flaml}, the design of meta-features for complex tasks in PD1 is non-trivial. In contrast, the text embedding approach has the additional advantage of not requiring any manual design for new types of tasks. Furthermore, text embedding is more promising for handling new and varied tasks, while meta-feature for new tasks is not easily scalable.

Nevertheless, we would like to emphasize that the key factor is not solely the method of retrieving experience, but rather how the retrieved experience is utilized. When the retrieved experience is used directly, as done in ASKL, all retrieval strategies perform poorly. In contrast, \name has the ability to not only retrieve relevant experience, but also provide guidance through elicited knowledge and leverage the power of LLMs.

\noindent \textbf{Study of canonicalization.} As shown in \autoref{tab:ablation-discretization}, the performance suffers considerably without discretization as sending continuous numerical values directly to LLM is not feasible. Furthermore, it is crucial to compute the split points based on the statistics of the best solutions. If the range is expanded to include all possible values, the split point may not fall on the sensitive points, resulting in subpar performance. This is demonstrated by ``On All'' in \autoref{tab:ablation-discretization}, which performs even worse than no discretization at all\footnote{Please note that HyperFD is not included in the ablation study as it adopts a discrete solution space.}.

\begin{table}[t]
\small
\centering
        \renewcommand{\tabcolsep}{0.7mm}

\begin{tabular}{cccc}
\toprule
\textbf{Retrieve}  & \textbf{HPO-B} $\uparrow$ & \textbf{PD1} $\uparrow$  & \textbf{HyperFD} $\downarrow$ \\ \midrule
  {Exp.+Know.} &   \meanstd{81.59}{0.94}  & \meanstd{1.48}{0.06} & \meanstd{59.74}{1.89} \\ 
{Know.}& \meanstd{62.77}{1.78} & \meanstd{1.10}{0.00} & \meanstd{127.75}{0.00} \\ 
  {Exp.} & \meanstd{76.21}{0.16} & \meanstd{1.36}{0.09} & \meanstd{63.20}{3.55}\\ 
 \bottomrule
\end{tabular}
\vspace{-0.1in}
\caption{Effect of retrieving experience and knowledge.}
\label{tab:ablation-experience}
\vspace{-0.1in}
\end{table}

\begin{table}[t]
\small
\centering
\renewcommand{\tabcolsep}{0.7mm}

\begin{tabular}{cccc}
\toprule
  {\textbf{Method}} 
 & {\textbf{HPO-B} $\uparrow$} & {\textbf{HyperBO} $\uparrow$} & {\textbf{HyperFD} $\downarrow$} \\ \midrule
\name & \meanstd{81.59}{0.94} & \meanstd{1.48}{0.06} & \meanstd{59.74}{1.89} \\  
\textbf{w/o} Post-Val.  & \meanstd{78.34}{0.71} & \meanstd{1.44}{0.05} & \meanstd{62.41}{3.66} \\ 
\textbf{w/o} Know.  & \meanstd{76.21}{0.16} & \meanstd{1.36}{0.09} & \meanstd{63.20}{3.55}\\ 

 \bottomrule
\end{tabular}
\vspace{-0.1in}
\caption{Ablation on knowledge utilization in online stage and post-validation in offline stage. }
\label{tab:ablation-knowledge}

\end{table}

\begin{table}[t]
\small
    \centering
        \renewcommand{\tabcolsep}{0.7mm}

\begin{tabular}{cccc}
\toprule

{\textbf{Discretization}} &  {\textbf{HPO-B}}$\uparrow$ & {\textbf{PD1}} $\uparrow$ & {\textbf{HyperFD}}$\downarrow$  \\ \midrule
On Best & \meanstd{81.59}{0.94} & \meanstd{1.48}{0.06} & \meanstd{59.74}{1.89} \\ 
On All & \meanstd{70.82}{0.01}  & \meanstd{1.41}{0.02} & -- \\ 
\xmark  &  \meanstd{74.09}{0.44}  & \meanstd{1.45}{0.05} & \meanstd{84.96}{7.16} \\ 
 \bottomrule
\end{tabular}
\vspace{-0.1in}

\caption{Discretization in canonicalization.}
\label{tab:ablation-discretization}

\vspace{-0.1in}
\end{table}

\noindent \textbf{Study of knowledge. }
In \autoref{tab:ablation-knowledge}, we conducted an ablation study to evaluate the impact of knowledge on our method. The results show that, removing knowledge retrieval in the online stage of our method results in a significant decrease in the final performance. This is because knowledge is instrumental in helping LLMs arrive at the most effective solutions. Furthermore, the post-validation in elicitation procedure in the offline stage of \name also plays a vital role in enhancing its usefulness. 

Based on our qualitative study on the generated knowledge (see \S~\ref{sec:appendix-knowledge}), we found that, the knowledge serves as a helpful summary of past ML experiences while providing guidance on how to adjust parameters and settings based on task characteristics.
We observe that post-validation significantly reduces the chances that trivial, vague, or hallucinated knowledge is produced, although such knowledge is still sometimes observed.
For example, in the case of ``UniRef50'' task with ``Transformer'' model on PD1, the knowledge contains certain numerical examples that were not part of the demonstrations and instead the result of hallucinations.

\section{Conclusion}

In conclusion, this paper proposes \name, a framework that unleashes the power of LLMs to solve practical ML tasks.
\name showcases the versitility of LLMs, that it can handle not only text-related tasks, but also tasks involving heterogeneous inputs and intricate reasoning. We believe this represents a significant advancement in expanding the scope of LLM applications to a broader spectrum of complex problems.

\section{Ethical considerations} 

The architecture of \name is meticulously engineered to ensure that the solutions it recommends always remain within the bounds of the solution space provided by the user. As a result, it acts as a safeguard against the generation of unethical solutions, provided that the defined solution space adheres to ethical standards.

However, the foundational techniques outlined in this paper, including experience retrieval and knowledge elicitation, possess broader applicability across various scenarios beyond machine learning, such as task automation~\cite{lu2023chameleon} and scientific research~\cite{boiko2023emergent}.
In these contexts where the solution space extends beyond the constraints of a strictly-defined machine learning problem and where Large Language Models (LLMs) exhibit inherent limitations, the potential for unpredictability arises. Therefore, it becomes imperative to exercise ethical prudence when deploying \name in diverse cases.

\section{Limitations}

\noindent \textbf{Potential data leakage.} Since LLMs are trained on large corpus of data from Internet, it is likely that the benchmarks (especially HPO-B based on OpenML) have already been encountered during the pre-training phase of LLMs. To mitigate this potential bias, we conducted an evaluation of \name on HyperFD~\cite{yan2022privacy}. It is worth noting that the HyperFD dataset was introduced in a paper published after the knowledge cutoff date of GPT-3.5, and the dataset itself remains private. We empirically reveal that \name exhibits robust performance on the HyperFD dataset.

Furthermore, our findings indicate a significant performance enhancement when the data is canonicalized (\autoref{tab:ablation-discretization}). If the data were indeed memorized during the pre-training process, LLMs would likely benefit from access to unaltered, raw data. These results provide valuable supporting evidence for the assertion that the capabilities of LLMs extend beyond mere memorization. They encompass a broader spectrum of cognitive skills, including mathematical reasoning and logical thinking.

\noindent \textbf{Distinction from AutoML methods.} \name is not intended to serve as a replacement for established AutoML approaches. The distinction is grounded in the inherent limitations of Large Language Models (LLMs) when it comes to performing mathematical computations, as illustrated in recent work~\cite{imani2023mathprompter}. Consequently, it is improbable that \name would surpass state-of-the-art Bayesian optimization methods in the pursuit of superior solutions. In \autoref{tab:performance-all} we terminated our evaluation at $t = 3$ (\emph{i.e.,} three solutions), as we observed that performance reached a point of saturation with further increases in $t$. 

We argue that the true value of \name lies in the following facets: \emph{(i)} it accepts arbitrary types of task descriptions; \emph{(ii)} it leverages ML experiences from diverse sources, encompassing both pretraining and prompting; \emph{(iii)} it exhibits an exceptional ability to rapidly produces multiple out-of-the-box solutions for a novel task. Consequently, we envision the possibility of combining \name with existing AutoML methods, opening up an intriguing avenue for future exploration.

\noindent \textbf{Robustness of \name.} As \name has the ability to accommodate heterogeneous formats of inputs, it is worth discussing the robustness of \name in the wild. This consideration extends to situations where users submit poorly-formatted task descriptions and when the experience pool includes data with noisy accuracy labels or flawed canonicalization. A detailed assessment of \name's robustness is presented in \S~\ref{sec:robustness}.

The experiments conducted shed light on the system's robustness against certain challenges (\emph{e.g.,} the choice of LLMs and task description formats). But it is still important to note that its performance can degrade under specific conditions, such as when dealing with a severely limited prompt context window length.

\medskip

\bibliography{main}

\clearpage

\onecolumn

\appendix

\section{Algorithms}
\label{sec:algorithms}

We summarize our method as \autoref{alg:offline} and \autoref{alg:online}.

\label{sec:appedix-alg}
\begin{algorithm}[h]
\SetKw{return}{return}
\SetKwFunction{Suggest}{Suggest}
\SetKwFunction{GenKnowledge}{GenKnowledge}
\SetKwFunction{Sample}{RandomSample}
\SetKwFunction{Evaluate}{Evaluate}
\SetKwFunction{Uniform}{Uniform}
\SetKwInOut{Input}{Input}
\SetKwInOut{Output}{Output}

	\Input{Historical data $H = \{D_1, \ldots, D_{N_H}\}$. Maximum iterations \textit{rounds}. Stagnation patience \textit{patience}. Candidate question list \textit{Questions}. Validation tasks \textit{ValTasks}.
 }
	\Output{Experience Pool $P_E$; Knowledge  $k^\ast$. }
        $P_E \leftarrow \{\mathcal{C}(D_i)\}$ \tcc*[Python]{$\mathcal{C}$ is canonicalization function}
        $r^\ast \leftarrow - \infty$;\\
        $\textit{stagnation} \leftarrow 0$; \\
        \For{n=1 \KwTo rounds} {
            $E \leftarrow$ \Sample{$P_E$};\\
            $q \leftarrow $ \Sample{Questions} \tcc*[Python]{Sample one hypophora question}
            $\tau \leftarrow $\Uniform{$0, 1$}; \tcc*[Python]{Random temperature}
            $k \leftarrow  \textsc{Llm}\left(E, q; \tau\right)$ \tcc*[Python]{ Generate knowledge candidates}
            $S \leftarrow \textsc{Llm}\left(E, k; \textit{ValTasks}; 0\right)$  \tcc*[Python]{ Mock online stage on validation tasks}
            $r \leftarrow \Evaluate(S)$ \tcc*{Run and evaluate the solution}

                    \If{$r > r^\ast$}{
                             $r\ast \leftarrow r$;\\
                             $k^\ast \leftarrow k$;\\
                             $\textit{stagnation} \leftarrow 0$; \\
                    }
                    \Else{
                        $\textit{stagnation} \leftarrow \textit{stagnation} + 1$; \\
                        \If{\textit{stagnation} > \textit{patience}}{
                            \textbf{break};
                        }
                    }
        }
        \return$P_E, \,k^\ast$\\
 	 \caption{Offline Stage of \name} 
   \label{alg:offline}
 	 \end{algorithm}

\begin{algorithm}[h]
\SetKw{return}{return}
\SetKwFunction{Suggest}{Suggest}
\SetKwFunction{GenKnowledge}{GenKnowledge}
\SetKwFunction{Sample}{RandomSample}
\SetKwFunction{Evaluate}{Evaluate}
\SetKwFunction{Uniform}{Uniform}
\SetKwInOut{Input}{Input}
\SetKwInOut{Output}{Output}

	\Input{A new task description $\tilde{T}$. Experience pool $P_E$. Knowledge pool $P_K$.
 }
	\Output{Solution $\tilde{S}$.}
        $\tilde{E} \leftarrow \mathcal{R}_E(\tilde{T}, P_E)$ \tcc*[Python]{Retrieve experiences}
        $\tilde{K} \leftarrow \mathcal{R}_K(\tilde{T}, P_K)$\tcc*[Python]{Retrieve knowledge}
        $\tilde{S} \leftarrow \textsc{Llm}(\tilde{T}, \tilde{E}, \tilde{K})$;\\
        \return $\tilde{S}$\\
 	 \caption{Online Stage of \name} 
   \label{alg:online}
 	 \end{algorithm}

\section{Experiment Details}

\subsection{Benchmarks}
\label{sec:appendix-benchmark}

\textbf{HPO-B.} HPO-B-v3~\cite{arango2hpo} comprises 16 solution spaces for 101 datasets obtained from the OpenML~\cite{vanschoren2014openml}. Each space has a fixed ML algorithm such as random forest~\cite{breiman2001random}, SVM~\cite{cortes1995support}, or XGBoost~\cite{chen2016xgboost}, and the goal is to determine the optimal configuration of the algorithm for a given dataset. HPO-B also provides successful configurations from past tasks, which are canonicalized into experience in our case. Additionally, they have released surrogate models to expedite the evaluation of solutions that have not been attempted before. The final benchmark performance is determined by averaging the normalized accuracy (nAcc) across all datasets, following the normalization protocol in \cite{arango2hpo}.

\textbf{PD1.} The PD1 Neural Net Tuning Dataset is proposed by HyperBO~\cite{wang2021pre}, consisting of $24$ classification tasks, covering image classification, next token prediction, and translation. 
Each task is associated with a predefined neural network (CNN~\cite{krizhevsky2017imagenet} or transformer~\cite{vaswani2017attention}), and has four configurable parameters of a SGD optimizer with Nesterov momentum~\cite{nesterov1983method}.
Due to the high cost of training neural networks, evaluating the solutions suggested by \name by running them in real-time is not feasible. 
So we created a surrogate model to predict the performance of suggested solutions for each task (see \S~\ref{sec:appendix-surrogate} for details). As per \cite{wang2021pre}, we report normalized accuracy (nAcc).

\textbf{HyperFD.} HyperFD~\cite{yan2022privacy} is a benchmark designed to optimize the performance of a neural face detector on an unseen dataset by properly configuring data augmentation, neural architecture, loss function, and training recipe. For this purpose, they formulated a solution space and provided both average precision (AP) and rank for every possible solution on all datasets. The benchmark was published after the claimed knowledge cutoff of GPT-3.5 (September 2021), and it has not been publicly released yet, making it unlikely that it has appeared in the training data of LLM.

\subsection{Compared Baselines}
\label{sec:appedix-baseline}

Details about the traditional baselines used in our experiment are described below.

\begin{itemize}[leftmargin=1em]
    \item \textbf{Random} method randomly generates a solution.
    \item \textbf{ASKL} (Auto-sklearn 1.0)~\cite{feurer2015efficient} finds the most similar tasks based on manually selected meta-features of tasks and directly uses the best solutions on them.
    \item \textbf{Constant}~\cite{bardenet2013collaborative, kotthoff2019auto} (a.k.a. Average) uses a constant set of solutions for any new task. The produced set of solutions is the one with the best \textbf{average} performance on the historical tasks. This method is straightforward and has no specific literature reference. We reference the two literatures for ``Constant'' as they also adopt a similar baseline for comparison.
    \item \textbf{TST-M}~\cite{wistuba2016two} employs Gaussian processes to approximate the performance of solutions in the solution space for each task. When a new task is encountered, it combines performance predictions of solutions for different tasks and predicts the performance of each solution on the new task by averaging predictions on history tasks weighted by task similarities.
    \item \textbf{HyperSTAR}~\cite{mittal2020hyperstar} trains a performance predictor for a joint encoding of solution and task features.
    HyperSTAR is originally built for vision tasks. To adapt it for non-image tasks, we incorporate handcrafted meta-features as task features.
    \item \textbf{HyperFD}~\cite{yan2022privacy} is a method specifically designed for the HyperFD benchmark, which uses a sophisticated meta-feature extractor for neural face detection tasks. However, it is not a general-purpose method and is not designed to work with other types of tasks.
    \item \textbf{FLAML-Zero}~\cite{wang2021flaml} is a recent method that generates a portfolio of ML solutions through offline meta-training, minimizing overall regret across meta-training tasks. It uses meta-features to link new tasks to existing ones based on their similarity.
\end{itemize}

\subsection{Post-validation}

The post-validation step that we have incorporated draws inspiration from established practices within machine learning, where a dedicated validation set is employed to enhance model performance. In our specific case, we allocate 10\% of the training meta-dataset for validation purposes, allowing us to systematically filter and select the most valuable generated knowledge. This additional layer of validation contributes significantly to adcressing hallucination-related issues.

Detailed steps of post-validation include a loop of: sampling hypophora question, sampling temperature, generating knowledge candidates, validating candidate knowledge, and an earlystopping mechanism that determines stagnation. This is described in \autoref{alg:offline}.

\subsection{Building Surrogate Model for PD1}
\label{sec:appendix-surrogate}

The metric in PD1 contains many NaN values, which correspond to network training divergence. For benchmarking purposes, it is more important to be able to distinguish the top-performing solutions, \emph{i.e.,} solutions above medium accuracy. To accomplish this, we adopt a two-stage surrogate approach. We use a classification model to distinguish the top-performing solutions, and then two regression models: one specially optimized for the top-performing solutions, and the other one for all solutions. We utilize XGBoost~\cite{chen2016xgboost} for building classifiers and regressors. Default parameters are used for those models.

\section{Prompt Design}
\label{sec:appendix-prompt-design}

We show two example prompts used on HPO-B. One of them is used for the online stage, as shown in \autoref{tab:prompt}, when \name receives a task description given by the user and sends it to LLM to obtain a recommended solution.

\autoref{tab:prompt-guidelines} shows an example of prompt used during the offline stage, when we generate a series of knowledge candidates. For post-validation, we use the prompt same as the online stage (example in \autoref{tab:prompt}).

\begin{table}[h]
\centering
\renewcommand{\arraystretch}{1.2}
\begin{minipage}{.57\textwidth}
    \centering
    \small

    \begin{tabular}{|m{.25\textwidth}m{.65\textwidth}|}
        \hline
        \multicolumn{2}{|C{.9\textwidth}|}{\textbf{Prompt}} \\ [2pt]
        \hline
        \multicolumn{2}{|C{.9\textwidth}|}{\textbf{Space description}} \\ [2pt]
       \cline{1-2}
        \multicolumn{2}{|L{.9\textwidth}|}{Here are some classification datasets along with best hyper-parameter configurations to train a R language model "Learner mlr.classif.svm from package(s) e1071" on them.} \\
       \cline{1-2}
        \multicolumn{2}{|C{.9\textwidth}|}{\textbf{Demonstrations}} \\
       \cline{1-2}
        \multicolumn{2}{|L{.9\textwidth}|}{Dataset: The dataset name is "ada\_agnostic". It contains 2 classes, 4562 instances, 49 features, 48 numeric features, 1 categorical features. The majority class size is 3430 and the minority class size is 1132.}\\
        \multicolumn{2}{|L{.9\textwidth}|}{Configuration 1: cost is very small. kernel is linear.}\\
        \multicolumn{2}{|L{.9\textwidth}|}{Configuration 2: cost is very small. kernel is linear.}\\
        \multicolumn{2}{|L{.9\textwidth}|}{Configuration 3: cost is very small. kernel is linear.} \\ \multicolumn{2}{|L{.9\textwidth}|}{} \\

        \multicolumn{2}{|L{.9\textwidth}|}{Dataset: The dataset name is "credit-g". It contains 2 classes, 1000 instances, 21 features, 7 numeric features, 14 categorical features. The majority class size is 700 and the minority class size is 300.}\\
        \multicolumn{2}{|L{.9\textwidth}|}{Configuration 1: cost is medium. gamma is small. kernel is radial.}\\
        \multicolumn{2}{|L{.9\textwidth}|}{Configuration 2: cost is medium. gamma is very small. kernel is radial.}\\
        \multicolumn{2}{|L{.9\textwidth}|}{Configuration 3: cost is medium. gamma is small. kernel is radial.} \\ \multicolumn{2}{|L{.9\textwidth}|}{}\\

        \multicolumn{2}{|L{.9\textwidth}|}{Dataset: The dataset name is "ozone-level-8hr". It contains 2 classes, 2534 instances, 73 features, 72 numeric features, 1 categorical features. The majority class size is 2374 and the minority class size is 160. }\\
        \multicolumn{2}{|L{.9\textwidth}|}{Configuration 1: cost is small. gamma is small. kernel is radial.}\\
        \multicolumn{2}{|L{.9\textwidth}|}{Configuration 2: cost is very small. gamma is small. kernel is radial.}\\
        \multicolumn{2}{|L{.9\textwidth}|}{Configuration 3: cost is small. gamma is small. kernel is radial. } \\
        
        \hline
        \multicolumn{2}{|C{.9\textwidth}|}{\textbf{Knowledge}} \\[1pt]
       \cline{1-2}
       \multicolumn{2}{|L{.9\textwidth}|}{Guidelines:} \\
        \multicolumn{2}{|L{.9\textwidth}|}{1. For datasets with many numeric features, larger cost values and smaller gamma values tend to be more effective.}\\
        \multicolumn{2}{|L{.9\textwidth}|}{2. For datasets with many categorical features, linear kernels tend to be more effective.}\\
        \multicolumn{2}{|L{.9\textwidth}|}{3. For datasets with few numeric features, small cost values and larger gamma values tend to be more effective.}\\
       \multicolumn{2}{|L{.9\textwidth}|}{4. For datasets with few categorical features, polynomial kernels tend to be more effective.} \\
        \cline{1-2}
         \multicolumn{2}{|C{.9\textwidth}|}{\textbf{Instruction}} \\
        \cline{1-2}
        \multicolumn{2}{|L{.9\textwidth}|}{Based on the examples and guidelines above, recommend 3 hyper-parameter configurations for a new classification dataset}\\
        \hline
        \multicolumn{2}{|C{.9\textwidth}|}{\textbf{Description for new task}} \\ [2pt]
       \cline{1-2}
        \multicolumn{2}{|L{.9\textwidth}|}{Dataset: The dataset name is "gina\_agnostic". It contains 2 classes, 3468 instances, 971 features, 970 numeric features, 1 categorical features. The majority class size is 1763 and the minority class size is 1705. } \\
       \hline
    \end{tabular}
        \caption{Example prompt for HPO-B in online serving.}
    \label{tab:prompt}

\end{minipage}%
\begin{minipage}{.43\textwidth}

    \centering
    \small

    \begin{tabular}{|m{.25\textwidth}m{.65\textwidth}|}
        \hline
        \multicolumn{2}{|C{.9\textwidth}|}{\textbf{Prompt}} \\ [2pt]
        \hline
        \multicolumn{2}{|C{.9\textwidth}|}{\textbf{Space description}} \\ [2pt]
       \cline{1-2}
        \multicolumn{2}{|L{.9\textwidth}|}{Here are some classification datasets along with best hyper-parameter configurations to train a R language model "Learner mlr.classif.svm from package(s) e1071" on them.} \\
       \cline{1-2}
        \multicolumn{2}{|C{.9\textwidth}|}{\textbf{Demonstrations}} \\
       \cline{1-2}
        \multicolumn{2}{|L{.9\textwidth}|}{Dataset: The dataset name is "wilt". It contains 2 classes, 4839 instances, 6 features, 5 numeric features, 1 categorical features. The majority class size is 4578 and the minority class size is 261.}\\
        \multicolumn{2}{|L{.9\textwidth}|}{Configuration 1: cost is medium. gamma is large. kernel is radial.}\\
        \multicolumn{2}{|L{.9\textwidth}|}{Configuration 2: cost is medium. gamma is medium. kernel is radial.}\\
        \multicolumn{2}{|L{.9\textwidth}|}{Configuration 3: cost is large. gamma is medium. kernel is radial.} \\ \multicolumn{2}{|L{.9\textwidth}|}{} \\

        \multicolumn{2}{|L{.9\textwidth}|}{Dataset: The dataset name is "ilpd". It contains 2 classes, 583 instances, 11 features, 9 numeric features, 2 categorical features. The majority class size is 416 and the minority class size is 167.}\\
        \multicolumn{2}{|L{.9\textwidth}|}{Configuration 1: cost is medium. gamma is medium. kernel is radial.}\\
        \multicolumn{2}{|L{.9\textwidth}|}{Configuration 2: cost is very small. gamma is very large. kernel is radial.}\\
        \multicolumn{2}{|L{.9\textwidth}|}{Configuration 3: cost is medium. gamma is very large. kernel is radial.} \\ \multicolumn{2}{|L{.9\textwidth}|}{}\\

        \multicolumn{2}{|L{.9\textwidth}|}{Dataset: The dataset name is "steel-plates-fault". It contains 2 classes, 1941 instances, 34 features, 33 numeric features, 1 categorical features. The majority class size is 1268 and the minority class size is 673.}\\
        \multicolumn{2}{|L{.9\textwidth}|}{Configuration 1: cost is small. kernel is linear.}\\
        \multicolumn{2}{|L{.9\textwidth}|}{Configuration 2: cost is very small. kernel is linear.}\\
        \multicolumn{2}{|L{.9\textwidth}|}{Configuration 3: cost is very small. kernel is linear. } \\
        
        \hline
        \multicolumn{2}{|C{.9\textwidth}|}{\textbf{Instruction}} \\[2pt]
       \cline{1-2}
        \multicolumn{2}{|L{.9\textwidth}|}{Q: From the examples above, what patterns can we observe about the relationship between dataset characteristics and the best hyper-parameter configurations? Answer MUST be concise, critical, point-by-point, line-by-line, and brief. Only include relevant observations without unnecessary elaboration.}\\
       \hline
    \end{tabular}
        \caption{Example prompt for HPO-B in the offline stage.}
    
    \label{tab:prompt-guidelines}
\end{minipage}
\end{table}

\clearpage

\section{Robustness of \name}
\label{sec:robustness}

\begin{table}[t]
\small
\centering
\caption{Comparison of different description formats. }
\label{tab:ablation-description}
\begin{tabular}{ccccc}
\toprule
 \textbf{Description Format} & {\textbf{HPO-B} $\uparrow$} & {\textbf{PD1} $\uparrow$} & {\textbf{HyperFD} $\downarrow$} \\  \midrule
Original & \meanstd{81.59}{0.94} & \meanstd{1.48}{0.06} & \meanstd{59.74}{1.89} \\  
Condense  & \meanstd{79.66}{0.06} & \meanstd{1.52}{0.01} & \meanstd{57.33}{3.21} \\ 
Anonymous  & \meanstd{77.43}{0.04} & \meanstd{1.21}{0.06} & \meanstd{68.42}{10.01} \\ 
\midrule
Misleading names  & \meanstd{75.80}{3.01} & \meanstd{1.43}{0.05} & \meanstd{62.52}{3.92} \\ 

 \bottomrule
\end{tabular}

\end{table}

\begin{figure}[t]
\centering
\begin{subfigure}[b]{0.31\textwidth}
   \includegraphics[width=\linewidth]{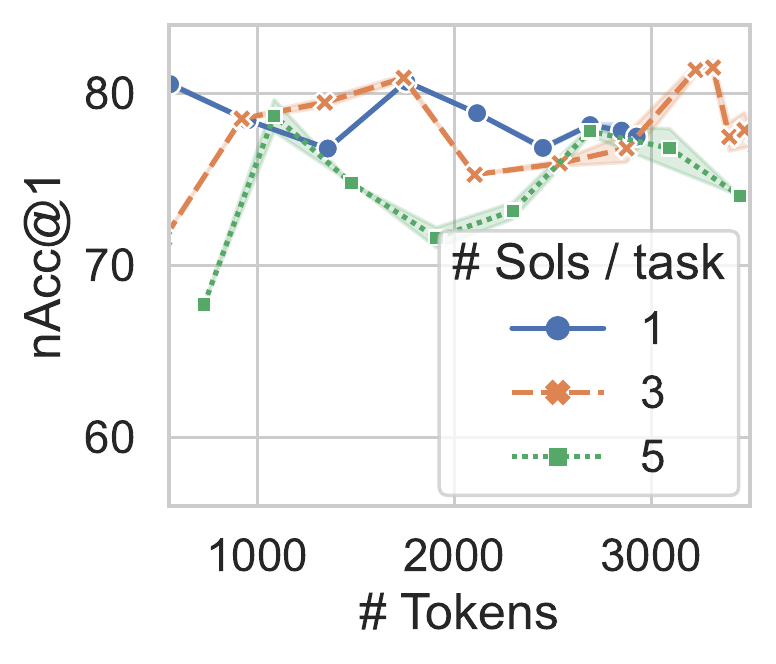}
   \vspace{-5mm}
   \caption{HPO-B (higher better).}
\end{subfigure}
\begin{subfigure}[b]{0.31\textwidth}
   \includegraphics[width=\linewidth]{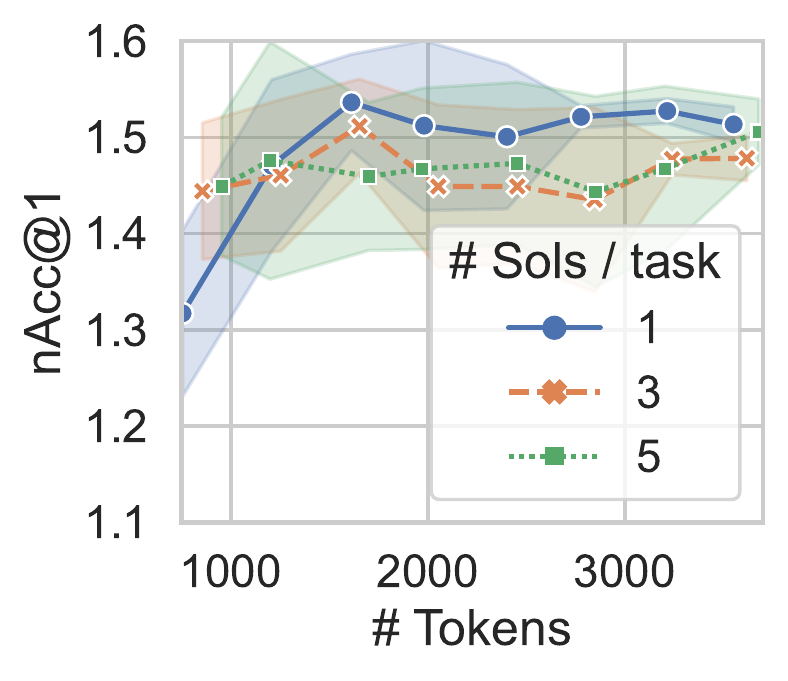}
   \vspace{-5mm}
   \caption{PD1 (higher better).}
\end{subfigure}
\begin{subfigure}[b]{0.31\textwidth}
   \includegraphics[width=\linewidth]{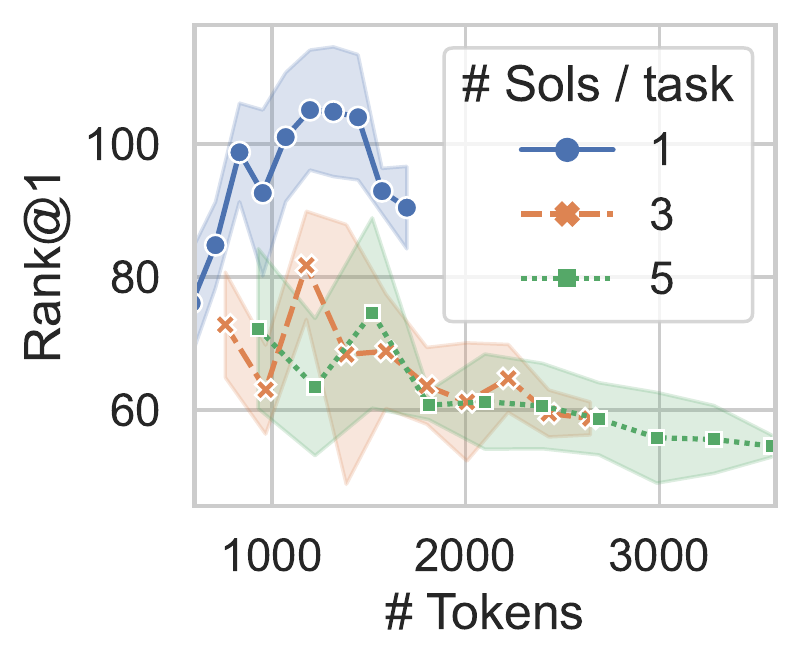}
   \vspace{-5mm}
   \caption{HyperFD (lower better).}
\end{subfigure}
\caption{Effect of the experience number and the number of solutions demonstrated for each task.}\label{fig:ablation-fewshot}
\vspace{-3mm}
\end{figure}

\subsection{Task description in the wild. }
\label{sec:task-description-in-the-wild}
When \name is deployed, it may be impractical to require users to strictly adhere to a specific format when writing task descriptions. We must consider if \name is robust enough to handle various formats. To simulate diverse formats, we ask GPT-3.5 to rewrite the descriptions by: \emph{(i)} condensing the original task descriptions; and \emph{(ii)} anonymizing the descriptions by removing task names.
The results are shown in \autoref{tab:ablation-description}. We observed fluctuations in performance when the description format changed, indicating that LLM is sensitive to prompt format. 
This aligns with the previous researches~\cite{webson2022prompt,lu2022fantastically} suggesting that LLMs may not interpret inputs in the same way as humans.
Performance was particularly worse when tasks were anonymized, leading us to conjecture that task names stimulate neurons in the LLM that are important for solving the relevant tasks and also leverage the previous knowledge already memorized from the training corpus of LLMs. 
To further verify this, we conducted an additional experiment by randomly swapping task names between tasks, and surprisingly observed performance improvement on PD1 and HyperFD. This echoes the finding in \cite{min2022rethinking}, which suggests that ``random labels are better than no labels at all''.

\subsection{Length of prompt.}
We study the effect of prompt length, which is mainly influenced by the number of retrieved ML experiences as demonstrations, to the performance.
In our previous experiments, we retrieved as many experiences as possible, either until all the experiences had been retrieved or the maximum prompt length was reached. 
For each task, the three best solutions were demonstrated, which was an arbitrary choice in our early experiments. 
In this section, we vary these two factors.
As shown in \autoref{fig:ablation-fewshot}, the performance generally improves as the number of demonstrations in the prompt increases, measured by prompt tokens.
However, the performance soon saturates at around 1.5k tokens and fluctuates. 
Moreover, demonstrating more solutions for each task leverages more data and has higher potential, especially above 3k tokens.

\subsection{Choice of LLMs.}
\label{sec:choice-of-llms}

We report performance of MLCopilot if equipped with LLMs other than GPT-3.5 (code-named ``text-davinci-003'') used in our main experiments. The models we have experimented with include:

\begin{itemize}
\item GPT-3.5 Turbo\footnote{\url{https://openai.com/blog/introducing-chatgpt-and-whisper-apis}}: a cost-efficient version of GPT-3.5 that uses chat completion as its user interface.
\item GPT-3~\cite{brown2020language} (code-named ``text-davinci-001''): the original GPT-3 model trained without instruction finetuning.
\item LLAMA-7B~\cite{touvron2023llama}: a well-known open-source model with a large user community, with a relatively loose requirement of GPU memory.
\end{itemize}

We compare the results with the original results with text-davinci-003 (GPT-3.5) and our main baselines. As shown in Table \autoref{tab:different-llms}, \name is robust to choices of LLMs. It is compatible with all the LLMs we have tested and achieves competitive results under different settings. Also, we see a trend that when working with stronger and larger models, \name still achieves even better results.

\begin{table}[t]
\small
\centering
\begin{tabular}{ccccc}
\toprule
 \textbf{Method (with LLM)} & {\textbf{HPO-B} $\uparrow$} & {\textbf{PD1} $\uparrow$} & {\textbf{HyperFD} $\downarrow$} \\  \midrule
ASKL                         & \meanstd{77.01}{0.00} & \meanstd{1.26}{0.00} & \meanstd{92.58}{0.00}  \\
FLAML                        & \meanstd{77.84}{0.00} & \meanstd{1.28}{0.00} & \meanstd{66.42}{0.00}  \\
MLCopilot (gpt-3.5-turbo)    & \meanstd{81.67}{1.99}      &  \meanstd{1.37}{0.10}         & \meanstd{72.41}{10.48}           \\
MLCopilot (text-davinci-001) & \meanstd{82.13}{2.05}      & \meanstd{1.58}{0.04}         & \meanstd{71.25}{10.70}       \\
MLCopilot (LLAMA-7B)         & \meanstd{79.51}{0.57} & \meanstd{1.43}{0.08} & \meanstd{67.47}{5.60}  \\
MLCopilot (text-davinci-003) & \meanstd{81.59}{0.94} & \meanstd{1.48}{0.06} & \meanstd{59.74}{1.89} \\ 

 \bottomrule
\end{tabular}
\caption{Performance of \name equipped with different LLMs.}
\label{tab:different-llms}

\end{table}

\subsection{Noisy accuracy and faulty canonicalization.}

\begin{table}[h]
\begin{minipage}{.5\textwidth}
    \centering
    \small
    \begin{tabular}{ccc}
    \toprule
    \textbf{Method} & \textbf{w/ original data} & \textbf{w/ perturbed data} \\
    \midrule
    FLAML & \meanstd{77.84}{0.00} & \meanstd{73.53}{0.00} \\
    \name & \meanstd{81.59}{0.94} & \meanstd{78.54}{3.25} \\
    \bottomrule
    \end{tabular}
    \caption{Impact of noises in accuracy.}
    \label{tab:accuracy-noise}
\end{minipage}%
\begin{minipage}{.5\textwidth}
    \centering
    \small
    \begin{tabular}{cc}
    \toprule
    \textbf{Method} & \textbf{nAcc} \\
    \midrule
    \name (original) & \meanstd{81.59}{0.94} \\
    \name (faulty canonicalization) & \meanstd{77.11}{1.77} \\
    FLAML & \meanstd{77.84}{0.00} \\
    \bottomrule
    \end{tabular}
    \caption{Effect of faulty canonicalization.}
    \label{tab:faulty-canonicalization}
\end{minipage}
\end{table}

We discuss the cases where the experience pool is polluted during the operation of the system. We evaluated the robustness under such scenarios on HPO-B (the largest solution space).

Firstly, we assessed the impact of noises in accuracy, by perturbing the accuracies in historical data. We added Gaussian noise to the accuracy values. The standard deviation of the Gaussian noise is 10\% of the accuracy distribution. As a result of such disturbance, suboptimal configurations might pop up as the best configurations and serve as demonstrations in the prompt. After experimenting on HPO-B, we found (in \autoref{tab:accuracy-noise}) that \name does suffer from such disturbance (performance drop from 81.59 to 78.54). However, such a result is still competitive with the state-of-the-art baselines (FLAML 77.84). Moreover, if a similar disturbance was done to the input data of FLAML, its performance further drops to 73.53. Full results (top-1 normalized accuracy) are shown in the table below.

Secondly, we investigate the effect of faulty canonicalization. In our paper, we showed (in \autoref{tab:ablation-discretization}) that canonicalization is an important component of MLCopilot, and a misconfigured canonicalization can lead to degraded performance. Following your suggestions, we introduced random noise into the canonicalization process by replacing 10\% of the canonicalized data with random discrete values. That is, each parameter of the configuration has a 10\% probability to be replaced with a random choice from ``very low'', ``low'', ``medium'', ``high'', ``very high''. We show the results (top-1 normalized accuracy on HPO-B) in \autoref{tab:faulty-canonicalization}.

Although the result is still competitive with baseline FLAML, we can see that faulty canonicalization does lead to worse performance. Notably, the impact is even more severe than the setting of perturbed accuracy. We speculate that false canonicalization can be particularly misleading for the logical reasoning of large language models. We will include a discussion of these findings in our revision.

\clearpage

\section{Knowledge}
\label{sec:appendix-knowledge}

All contents in this section are generated by Large Language Models.

\subsection{HPO-B}

HPO-B contains 16 design spaces. We finalize one set of knowledge for each space.

\lstset{
  basicstyle=\ttfamily,
  columns=fullflexible,
  keepspaces=true,
  escapeinside=||
}

\paragraph{Space: 5860}
\vphantom{X}

\begin{lstlisting}
1. Generally, datasets with more |\textcolor{cyan}{numeric features}| require larger
   alphas and smaller lambdas for better performance.
2. Datasets with a higher ratio of minority to majority |\textcolor{cyan}{class size}|
   require smaller alphas and larger lambdas for better performance.
3. Datasets with |\textcolor{cyan}{more features}| require larger alphas and smaller
   lambdas for better performance.
4. Datasets with more |\textcolor{cyan}{categorical}| features require larger alphas and
   larger lambdas for better performance.
\end{lstlisting}

\paragraph{Space: 4796}
\vphantom{X}

\begin{lstlisting}
1. For datasets with a large majority |\textcolor{cyan}{class size}| and a small minority
   |\textcolor{cyan}{class size}|, a larger cp and minbucket size tend to be better
   hyper-parameter configurations.
2. For datasets with a small majority |\textcolor{cyan}{class size}| and a large minority
   |\textcolor{cyan}{class size}|, a smaller cp and minbucket size tend to be better
   hyper-parameter configurations.
3. For datasets with a large number of |\textcolor{cyan}{numeric features}|, a larger cp
   and minbucket size tend to be better hyper-parameter configurations.
4. For datasets with a small number of |\textcolor{cyan}{numeric features}|, a smaller cp
   and minbucket size tend to be better hyper-parameter configurations.
5. For datasets with a large number of |\textcolor{cyan}{categorical}| features, a
   smaller cp and minbucket size tend to be better hyper-parameter
   configurations.
6. For datasets with a small number of |\textcolor{cyan}{categorical}| features, a larger
   cp and minbucket size tend to be better hyper-parameter
   configurations.
\end{lstlisting}

\paragraph{Space: 5971}
\vphantom{X}

\begin{lstlisting}
1. Generally, |\textcolor{cyan}{larger datasets}| require higher nrounds and larger
   subsample values.
2. The majority |\textcolor{cyan}{class size}| and minority |\textcolor{cyan}{class size}| of the dataset can
   influence the configuration of alpha, booster, colsample bylevel,
   colsample bytree, eta, lambda, max depth, min child weight, nrounds,
   and subsample.
3. The number of numeric and |\textcolor{cyan}{categorical}| features in the dataset can
   determine the booster used.
4. The |\textcolor{cyan}{size of the dataset}| can influence the configuration of eta,
   lambda, max depth, min child weight, nrounds, and subsample.
5. The |\textcolor{cyan}{size of the minority}| class can determine the configuration of
   alpha, colsample bylevel, colsample bytree, eta, lambda, max depth,
   min child weight, nrounds, and subsample.
\end{lstlisting}

\paragraph{Space: 6766}
\vphantom{X}

\begin{lstlisting}
1. For datasets with a larger majority |\textcolor{cyan}{class size}|, high values of
   alpha and low values of lambda tend to perform better.
2. For datasets with a smaller majority |\textcolor{cyan}{class size}|, low values of
   alpha and high values of lambda tend to perform better.
3. For datasets with more |\textcolor{cyan}{numeric features}|, medium values of alpha
   and low values of lambda tend to perform better.
4. For datasets with more |\textcolor{cyan}{categorical}| features, high values of alpha
   and large values of lambda tend to perform better.
5. For datasets with a |\textcolor{cyan}{larger number of features}|, high values of
   alpha and large values of lambda tend to perform better.
\end{lstlisting}

\paragraph{Space: 5965}
\vphantom{X}

\begin{lstlisting}
1. The larger the majority |\textcolor{cyan}{class size}|, the smaller the min node size
   and sample fraction tend to be.
2. The larger the minority |\textcolor{cyan}{class size}|, the larger the min node size
   and sample fraction tend to be.
3. The |\textcolor{cyan}{larger the number of features}|, the larger the mtry tends to be.
4. The larger the number of |\textcolor{cyan}{numeric features}|, the larger the mtry
   tends to be.
5. The larger the number of |\textcolor{cyan}{categorical}| features, the smaller the
   mtry tends to be.
6. The |\textcolor{cyan}{larger the number of trees}|, the smaller the mtry tends to be.
7. The |\textcolor{cyan}{larger the number of instances}|, the larger the sample fraction
   tends to be.
8. The replace parameter is |\textcolor{cyan}{usually}| set to True.
9. The respect unordered factors parameter is |\textcolor{cyan}{usually}| set to False.
\end{lstlisting}

\paragraph{Space: 5906}
\vphantom{X}

\begin{lstlisting}
1. |\textcolor{cyan}{Smaller datasets }| tend to have smaller alpha and eta values, while
   |\textcolor{cyan}{larger datasets}| tend to have larger values.
2. Datasets with |\textcolor{cyan}{more features}| tend to have larger colsample bylevel
   and colsample bytree values, while datasets with fewer features tend
   to have smaller values.
3. Datasets with more |\textcolor{cyan}{numeric features}| tend to have larger lambda and
   max depth values, while datasets with fewer |\textcolor{cyan}{numeric features}| tend to
   have smaller values.
4. |\textcolor{cyan}{Smaller datasets }| tend to have smaller nrounds and subsample
   values, while |\textcolor{cyan}{larger datasets}| tend to have larger values.
5. Datasets with more |\textcolor{cyan}{categorical}| features tend to have smaller min
   child weight values, while datasets with fewer |\textcolor{cyan}{categorical}| features
   tend to have larger values.
\end{lstlisting}

\paragraph{Space: 7607}
\vphantom{X}

\begin{lstlisting}
1. The min node size generally decreases as the |\textcolor{cyan}{dataset size
   increases}|.
2. The mtry is |\textcolor{cyan}{usually}| small for datasets with few features and large
   for datasets with many features.
3. The num trees is |\textcolor{cyan}{usually}| small for datasets with few instances and
   large for datasets with many instances.
4. Replace is |\textcolor{cyan}{usually}| set to False for small datasets and True for
   large datasets.
5. Respect unordered factors is |\textcolor{cyan}{usually}| set to False for datasets
   with few |\textcolor{cyan}{categorical}| features and True for datasets with many
   |\textcolor{cyan}{categorical}| features.
6. Sample fraction is |\textcolor{cyan}{usually}| set to small for datasets with few
   instances and large for datasets with many instances.
\end{lstlisting}

\paragraph{Space: 6794}
\vphantom{X}

\begin{lstlisting}
1. For datasets with a large majority |\textcolor{cyan}{class size}|, larger min node
   size and sample fraction values are |\textcolor{cyan}{usually}| used, while for datasets
   with a smaller majority |\textcolor{cyan}{class size}|, smaller min node size and sample
   fraction values are |\textcolor{cyan}{usually}| used.
2. For datasets with |\textcolor{cyan}{more features}|, larger mtry values are |\textcolor{cyan}{usually}|
   used.
3. For datasets with more |\textcolor{cyan}{numeric features}|, replace is |\textcolor{cyan}{usually}| set to
   True, while for datasets with more |\textcolor{cyan}{categorical}| features, replace is
   |\textcolor{cyan}{usually}| set to False.
4. Respect unordered factors is |\textcolor{cyan}{usually}| set to True when the dataset
   has more |\textcolor{cyan}{categorical}| features.
\end{lstlisting}

\paragraph{Space: 7609}
\vphantom{X}

\begin{lstlisting}
1. For datasets with |\textcolor{cyan}{more features}|, larger mtry values are preferred.
2. For datasets with |\textcolor{cyan}{more instances}|, larger sample fractions are
   preferred.
3. For datasets with |\textcolor{cyan}{more majority class instances}|, smaller min node
   sizes are preferred.
4. For datasets with more |\textcolor{cyan}{numeric features}|, replace is typically set
   to True.
5. For datasets with more |\textcolor{cyan}{categorical}| features, respect unordered
   factors is typically set to False.
6. For datasets with a more balanced |\textcolor{cyan}{class size}|, num trees is
   typically set to a smaller value.
\end{lstlisting}

\paragraph{Space: 5859}
\vphantom{X}

\begin{lstlisting}
1. |\textcolor{cyan}{larger datasets}| tend to require smaller cp values and larger
   minbucket values.
2. |\textcolor{cyan}{Smaller datasets }| tend to require larger cp values and smaller
   minbucket values.
3. For |\textcolor{cyan}{larger datasets}|, maxdepth tends to be very large or medium,
   whereas for |\textcolor{cyan}{Smaller datasets }|, maxdepth tends to be very small or
   small.
4. For |\textcolor{cyan}{larger datasets}|, minsplit tends to be very large or large,
   whereas for |\textcolor{cyan}{Smaller datasets }|, minsplit tends to be very small or
   small.
\end{lstlisting}

\paragraph{Space: 5889}
\vphantom{X}

\begin{lstlisting}
1. The |\textcolor{cyan}{larger the dataset size}|, the larger the mtry and num trees,
   and the smaller the sample fraction.
2. The larger the majority |\textcolor{cyan}{class size}|, the larger the mtry and num
   trees, and the smaller the sample fraction.
3. The |\textcolor{cyan}{smaller the number of features}|, the smaller the mtry and num
   trees, and the larger the sample fraction.
4. The more |\textcolor{cyan}{numeric features}|, the larger the mtry and num trees, and
   the smaller the sample fraction.
5. The more |\textcolor{cyan}{categorical}| features, the smaller the mtry and num trees,
   and the larger the sample fraction.
6. The replace parameter is |\textcolor{cyan}{usually}| set to True.
\end{lstlisting}

\paragraph{Space: 6767}
\vphantom{X}

\begin{lstlisting}
1. Datasets with a larger majority |\textcolor{cyan}{class size}| tend to require larger
   nrounds and larger subsample values.
2. Datasets with more |\textcolor{cyan}{numeric features}| tend to require larger
   colsample bylevel and colsample bytree values.
3. Datasets with more |\textcolor{cyan}{categorical}| features tend to require smaller
   min child weight values.
4. Datasets with a smaller minority |\textcolor{cyan}{class size}| tend to require
   smaller eta and lambda values.
5. Datasets with |\textcolor{cyan}{more features}| tend to require larger max depth
   values.
\end{lstlisting}

\paragraph{Space: 5970}
\vphantom{X}

\begin{lstlisting}
1. For datasets with more |\textcolor{cyan}{numeric features}|, smaller alpha and smaller
   lambda values tend to be the best hyper-parameter configurations.
2. For datasets with more |\textcolor{cyan}{categorical}| features, larger alpha and
   larger lambda values tend to be the best hyper-parameter
   configurations.
3. For datasets with majority |\textcolor{cyan}{class size}| significantly larger than
   minority |\textcolor{cyan}{class size}|, larger alpha and larger lambda values tend to be
   the best hyper-parameter configurations.
\end{lstlisting}

\paragraph{Space: 5527}
\vphantom{X}

\begin{lstlisting}
1. The cost parameter tends to increase as the |\textcolor{cyan}{dataset size increases}|.
2. The gamma parameter tends to decrease as the number of numeric
   features increases.
3. The kernel parameter tends to be radial for datasets with numeric
   features, and polynomial or linear for datasets with |\textcolor{cyan}{categorical}|
   features.
4. The degree parameter tends to increase as the number of
   |\textcolor{cyan}{categorical}| features increases.
\end{lstlisting}

\paragraph{Space: 5636}
\vphantom{X}

\begin{lstlisting}
1. The larger the majority |\textcolor{cyan}{class size}|, the smaller the cp value
   should be. 
2. The larger the minority |\textcolor{cyan}{class size}|, the larger the cp value should
   be.
3. The |\textcolor{cyan}{larger the number of features}|, the smaller the maxdepth value
   should be.
4. The larger the number of |\textcolor{cyan}{numeric features}|, the larger the
   minbucket value should be.
5. The larger the number of |\textcolor{cyan}{categorical}| features, the smaller the
   minbucket value should be.
6. The |\textcolor{cyan}{larger the number of instances}|, the larger the minsplit value
   should be.
\end{lstlisting}

\paragraph{Space: 5891}
\vphantom{X}

\begin{lstlisting}
1. For datasets with many |\textcolor{cyan}{numeric features}|, larger cost values and
   smaller gamma values tend to be more effective.
2. For datasets with many |\textcolor{cyan}{categorical}| features, linear kernels tend
   to be more effective.
3. For datasets with few |\textcolor{cyan}{numeric features}|, small cost values and
   larger gamma values tend to be more effective.
4. For datasets with few |\textcolor{cyan}{categorical}| features, polynomial kernels
   tend to be more effective.
\end{lstlisting}

\subsection{PD1}

We performed leave-one-out evaluation on the PD1 benchmark, which consists of 23 tasks. However, some tasks are using the same model and dataset but only different in batch size. These tasks should not appear in training tasks and test tasks at the same time~\cite{wang2021pre}. Therefore, only 13 distinct sets of training tasks were available for testing. For each set of training tasks, we generated a corresponding set of knowledge, which is presented below.

\paragraph{Test task: CIFAR100, Wide ResNet}
\vphantom{X}

\begin{lstlisting}
1. Set the initial learning rate (LR) according to the |\textcolor{cyan}{size of the}|
   |\textcolor{cyan}{dataset}| and the complexity of the model.
2. Set the momentum parameter to a lower value for |\textcolor{cyan}{larger datasets}|
   and a higher value for |\textcolor{cyan}{simpler models}|.
3. Set the power parameter to a higher value for |\textcolor{cyan}{more complex models}|.
4. Set the lambda parameter to a higher value for |\textcolor{cyan}{more complex models}|
   and a lower value for |\textcolor{cyan}{simpler models}|.
\end{lstlisting}

\paragraph{Test task: CIFAR10, Wide ResNet}
\vphantom{X}

\begin{lstlisting}
1. Adjust the initial learning rate and momentum based on the size
   and complexity of the dataset: higher for |\textcolor{cyan}{large and complex datasets}|,
   lower for |\textcolor{cyan}{small and simple datasets}|.
2. Adjust the power and lambda parameters based on the desired |\textcolor{cyan}{speed}|
   |\textcolor{cyan}{of the learning process}|: higher power and lower lambda for faster
   learning, lower power and higher lambda for slower learning.
3. Consider any |\textcolor{cyan}{domain-specific constraints}| when configuring the
   optimizer, |\textcolor{cyan}{such as}| accuracy requirements.
\end{lstlisting}

\paragraph{Test task: Fashion-MNIST, Max Pooling CNN with ReLU}
\vphantom{X}

\begin{lstlisting}
1. Set the initial learning rate to a |\textcolor{cyan}{low or medium}| value.
2. Set the momentum to a |\textcolor{cyan}{high or medium}| value.
3. Set the power to a |\textcolor{cyan}{low or medium}| value.
4. Set the lambda to a |\textcolor{cyan}{low or medium}| value.
5. Adjust the initial learning rate, momentum, power, and lambda
   according to the characteristics of the task, |\textcolor{cyan}{such as}| the |\textcolor{cyan}{dataset}|
   |\textcolor{cyan}{size, model architecture, and the complexity}| of the prediction task.
   For example, for tasks with |\textcolor{cyan}{larger datasets}|, a higher initial
   learning rate may be beneficial, while for tasks with smaller
   datasets, a lower initial learning rate may be more suitable.
   Similarly, for tasks with |\textcolor{cyan}{more complex models}|, a higher momentum may
   be beneficial, while for |\textcolor{cyan}{simpler models}|, a lower momentum may be more
   suitable. Additionally, for tasks with more |\textcolor{cyan}{complex prediction tasks}|,
   a higher power may be beneficial, while for |\textcolor{cyan}{simpler tasks}|, a lower
   power may be more suitable. Finally, for tasks with more complex
   models, a higher lambda may be beneficial, while for |\textcolor{cyan}{simpler models}|,
   a lower lambda may be more suitable.
\end{lstlisting}

\paragraph{Test task: Fashion-MNIST, Max Pooling CNN with Tanh}
\vphantom{X}

\begin{lstlisting}
1. Choose an initial LR that is appropriate for the |\textcolor{cyan}{size of the}|
   |\textcolor{cyan}{dataset}| and complexity of the model.
2. Set the momentum to a value that is appropriate for |\textcolor{cyan}{the size of}|
   |\textcolor{cyan}{the dataset}| and complexity of the model.
3. Set the power parameter to a value that is appropriate for the
   |\textcolor{cyan}{size of the dataset}| and complexity of the model.
4. Set the lambda parameter to a value that is appropriate for the
   |\textcolor{cyan}{size of the dataset}| and complexity of the model.
\end{lstlisting}

\paragraph{Test task: Fashion-MNIST, Simple CNN}
\vphantom{X}

\begin{lstlisting}
1. Set the initial learning rate (LR) to a value that is appropriate
   for the |\textcolor{cyan}{size of the dataset.}|
2. Set the momentum to a value that is appropriate for |\textcolor{cyan}{the size of}|
   |\textcolor{cyan}{the dataset}|.
3. Set the power parameter to a value that is appropriate for the
   |\textcolor{cyan}{size of the dataset.}|
4. Set the lambda parameter to a value that is appropriate for the
   |\textcolor{cyan}{size of the dataset}| and the desired |\textcolor{cyan}{level of regularization}|.
\end{lstlisting}

\paragraph{Test task: ImageNet, ResNet50}
\vphantom{X}

\begin{lstlisting}
1. For tasks with |\textcolor{cyan}{larger batch sizes}|, use a higher initial learning
   rate and higher momentum. For tasks with |\textcolor{cyan}{smaller batch size}|s, use a
   lower initial learning rate and lower momentum.
2. For tasks with |\textcolor{cyan}{ larger vocabularies}|, use a higher lambda value. For
   tasks with smaller vocabularies, use a lower lambda value.
3. For tasks with |\textcolor{cyan}{more complex models}|, use a higher power value. For
   tasks with |\textcolor{cyan}{simpler models}|, use a lower power value.
\end{lstlisting}

\paragraph{Test task: LM1B, Transformer}
\vphantom{X}

\begin{lstlisting}
1. Set the initial learning rate to a value that is suitable for the
   |\textcolor{cyan}{size and complexity of the dataset}|.
2. Set the momentum to a value that is suitable for the |\textcolor{cyan}{size and}|
   |\textcolor{cyan}{complexity of the dataset}|.
3. Set the power parameter to a value that is suitable for the |\textcolor{cyan}{noise}|
   |\textcolor{cyan}{and outliers in the dataset}|.
4. Set the lambda parameter to a value that is suitable for the |\textcolor{cyan}{noise}|
   |\textcolor{cyan}{and outliers in the dataset}|.
\end{lstlisting}

\paragraph{Test task: MNIST, Max Pooling CNN with ReLU}
\vphantom{X}

\begin{lstlisting}
1. Set the initial learning rate to a |\textcolor{cyan}{low or medium}| value.
2. Set the momentum to a |\textcolor{cyan}{high or medium}| value.
3. Set the power to a |\textcolor{cyan}{low or medium}| value.
4. Set the lambda to a low or high value.
5. Consider the characteristics of the task, |\textcolor{cyan}{such as}| the |\textcolor{cyan}{dataset}|
   |\textcolor{cyan}{size, model architecture, and the complexity}| of the prediction task,
   when adjusting the parameters.
6. For tasks with |\textcolor{cyan}{larger datasets}|, a higher initial learning rate and
   lower momentum may be more suitable.
7. For tasks with |\textcolor{cyan}{Smaller datasets }|, a lower initial learning rate and
   higher momentum may be more suitable.
8. For tasks with |\textcolor{cyan}{more complex models}|, a higher initial learning rate
   and lower momentum may be more suitable.
9. For tasks with |\textcolor{cyan}{simpler models}|, a lower initial learning rate and
   higher momentum may be more suitable.
10. For tasks with more |\textcolor{cyan}{complex prediction tasks}|, a higher initial
    learning rate and lower momentum may be more suitable.
11. For tasks with |\textcolor{cyan}{simpler prediction tasks}|, a lower initial learning
    rate and higher momentum may be more suitable.
\end{lstlisting}

\paragraph{Test task: MNIST, Max Pooling CNN with Tanh}
\vphantom{X}

\begin{lstlisting}
1. Set the initial learning rate to a high value to ensure that the
   model is able to learn quickly and efficiently.
2. Set the momentum to a low value to prevent the model from
   overfitting.
3. Set the power and/or lambda to high values to ensure that the
   learning rate decays slowly and the model is able to continue
   learning for a longer period of time.
4. For tasks |\textcolor{cyan}{such as}| training a CNN with max-pool and ReLU on Fashion
   MNIST, set the initial learning rate to a low value to prevent the
   model from overfitting.
5. For tasks |\textcolor{cyan}{such as}| training a ResNet50 on ImageNet, set the initial
   learning rate to a high value to ensure that the model is able to
   learn quickly and efficiently.
6. For tasks |\textcolor{cyan}{such as}| training a Wide ResNet on CIFAR100, set the
   initial learning rate to a very high value to ensure that the model
   is able to learn quickly and efficiently.
7. For tasks |\textcolor{cyan}{such as}| training a Transformer on UniRef50, set the
   initial learning rate to a low value to prevent the model from
   overfitting.
\end{lstlisting}

\paragraph{Test task: MNIST, Simple CNN}
\vphantom{X}

\begin{lstlisting}
1. Adjust the initial learning rate and momentum according to the
   |\textcolor{cyan}{size and complexity of the dataset}|: higher for large and complex
   datasets, lower for |\textcolor{cyan}{small and simple datasets}|.
2. Adjust the power and lambda parameters according to the |\textcolor{cyan}{size and}|
   |\textcolor{cyan}{complexity of the dataset}|: higher for |\textcolor{cyan}{large and complex datasets}|,
   lower for |\textcolor{cyan}{small and simple datasets}|.
3. Adjust the initial learning rate and momentum according to the
   task requirements: higher for tasks requiring high accuracy, lower
   for tasks requiring high speed.
\end{lstlisting}

\paragraph{Test task: SVHN, Wide ResNet}
\vphantom{X}

\begin{lstlisting}
1. For |\textcolor{cyan}{image classification tasks}|, set the initial learning rate (LR)
   to a higher value and the momentum to a lower value.
2. For |\textcolor{cyan}{language tasks}|, set the initial LR to a lower value and the
   momentum to a higher value.
3. For tasks with |\textcolor{cyan}{larger batch sizes}|, set the initial LR to a higher
   value and the momentum to a lower value.
4. For tasks with |\textcolor{cyan}{smaller batch size}|s, set the initial LR to a lower
   value and the momentum to a higher value.
5. For tasks with |\textcolor{cyan}{more complex models}|, set the power to a higher
   value and the lambda to a higher value.
6. For tasks with |\textcolor{cyan}{simpler models}|, set the power to a lower value and
   the lambda to a lower value.
\end{lstlisting}

\paragraph{Test task: UniRef50, Transformer}
\vphantom{X}

\begin{lstlisting}
1. Set the initial learning rate, momentum, power, and lambda values
   according to the following guidelines:
 - For |\textcolor{cyan}{larger batch sizes}| and |\textcolor{cyan}{larger datasets}|, use a higher learning
   rate, higher momentum, lower power, and higher lambda.
  - For |\textcolor{cyan}{smaller batch size}|s and |\textcolor{cyan}{Smaller datasets }|, use a lower
   learning rate, lower momentum, higher power, and lower lambda.
2. Examples:
 - For a CNN with max-pool and ReLU on Fashion MNIST with a batch
   size of 256, use an initial learning rate of 0.001, a momentum of
   0.9, a power of 0.1, and a lambda of 0.01.
 - For a Wide ResNet on CIFAR10 with a batch size of 2048, use an
   initial learning rate of 0.01, a momentum of 0.9, a power of 0.5, and
   a lambda of 0.001.
 - For a Transformer on LM1B with a batch size of 2048, use an
   initial learning rate of 0.001, a momentum of 0.9, a power of 0.01,
   and a lambda of 0.001.
\end{lstlisting}

\paragraph{Test task: WMT15, xformer}
\vphantom{X}

\begin{lstlisting}
1. Set the initial learning rate to a |\textcolor{cyan}{low or medium}| value.
2. Set the momentum to a |\textcolor{cyan}{high or medium}| value.
3. Set the power to a |\textcolor{cyan}{low or medium}| value.
4. Set the lambda to a |\textcolor{cyan}{high or medium}| value.
5. Adjust the initial learning rate and momentum based on the
   characteristics of the task, |\textcolor{cyan}{such as}| the dataset size, model
   architecture, and the complexity of the prediction task. For example,
   for tasks with |\textcolor{cyan}{larger datasets}|, a higher initial learning rate and a
   lower momentum may be more suitable, while for tasks with smaller
   datasets, a lower initial learning rate and a higher momentum may be
   more suitable. Additionally, for tasks with |\textcolor{cyan}{more complex models}|, a
   higher initial learning rate and a lower momentum may be more
   suitable, while for tasks with |\textcolor{cyan}{simpler models}|, a lower initial
   learning rate and a higher momentum may be more suitable. Finally,
   for tasks with more |\textcolor{cyan}{complex prediction tasks}|, a higher initial
   learning rate and a lower momentum may be more suitable, while for
   tasks with |\textcolor{cyan}{simpler prediction tasks}|, a lower initial learning rate
   and a higher momentum may be more suitable.
\end{lstlisting}

\subsection{HyperFD}

Similar to PD1, evaluation on HyperFD is also leave-one-out on 12 tasks. We show 12 sets of knowledge based on the choices of test tasks.

\paragraph{Test task: AFLW}
\vphantom{X}

\begin{lstlisting}
1. Configure crop size and anchor matching IoU threshold based on the
   |\textcolor{cyan}{number of faces in the dataset}|:
 - For |\textcolor{cyan}{datasets with more faces}|, use larger crop sizes and higher
   anchor matching IoU thresholds.
 - For |\textcolor{cyan}{datasets with fewer faces}|, use smaller crop sizes and lower
   anchor matching IoU thresholds.
2. Configure learning rate and negative to positive ratio based on
   the |\textcolor{cyan}{number of faces in the dataset}|:
 - For |\textcolor{cyan}{datasets with more faces}|, use higher learning rates and more
   negative to positive ratios.
 - For |\textcolor{cyan}{datasets with fewer faces}|, use lower learning rates and fewer
   negative to positive ratios.
3. Configure location loss weight based on the |\textcolor{cyan}{presence of facial}|
   |\textcolor{cyan}{landmarks in the dataset}|:
 - For datasets with |\textcolor{cyan}{facial landmarks}|, use higher location loss
   weights.
\end{lstlisting}

\paragraph{Test task: ANIME}
\vphantom{X}

\begin{lstlisting}
1. Set the crop size and anchor matching IoU threshold according to
   the |\textcolor{cyan}{number of faces in the dataset}|:
 - For |\textcolor{cyan}{datasets with more faces}|, use larger crop sizes and higher
   anchor matching IoU thresholds.
 - For |\textcolor{cyan}{datasets with fewer faces}|, use smaller crop sizes and lower
   anchor matching IoU thresholds.
2. Set the location loss weight according to the |\textcolor{cyan}{presence of facial}|
   |\textcolor{cyan}{landmarks in the dataset}|:
 - For datasets with |\textcolor{cyan}{facial landmarks}|, use higher location loss
   weights.
 - For datasets without |\textcolor{cyan}{facial landmarks}|, use lower location loss
   weights.
3. Set the learning rate and optimizer according to the |\textcolor{cyan}{negative to}|
   |\textcolor{cyan}{positive ratio in the dataset}|:
 - For datasets with higher negative to positive ratios, use higher
   learning rates and optimizers |\textcolor{cyan}{such as}| SGD or Adam.
\end{lstlisting}

\paragraph{Test task: FaceMask}
\vphantom{X}

\begin{lstlisting}
1. Set the crop size according to the |\textcolor{cyan}{number of faces in the dataset}|:
   larger crop sizes for |\textcolor{cyan}{datasets with more faces}|, and smaller crop
   sizes for |\textcolor{cyan}{datasets with fewer faces}|.
2. Set the anchor matching IoU threshold according to |\textcolor{cyan}{the number of}|
   |\textcolor{cyan}{faces}| in the dataset: higher thresholds for |\textcolor{cyan}{datasets with more faces}|,
   and lower thresholds for |\textcolor{cyan}{datasets with fewer faces}|.
3. Set the location loss weight according to the |\textcolor{cyan}{presence of facial}|
   |\textcolor{cyan}{landmarks in the dataset}|: higher weights for datasets with facial
   landmarks, and lower weights for datasets without |\textcolor{cyan}{facial landmarks}|.
4. Set the negative to positive ratio according to |\textcolor{cyan}{the number of}|
   |\textcolor{cyan}{faces}| in the dataset: higher ratios for |\textcolor{cyan}{datasets with more faces}|, and
   lower ratios for |\textcolor{cyan}{datasets with fewer faces}|.
5. Set the learning rate according to the number of faces in the
   dataset: higher rates for |\textcolor{cyan}{datasets with more faces}|, and lower rates
   for |\textcolor{cyan}{datasets with fewer faces}|.
\end{lstlisting}

\paragraph{Test task: FDDB}
\vphantom{X}

\begin{lstlisting}
1. Set the crop size to be larger and the anchor matching IoU
   threshold to be higher for |\textcolor{cyan}{datasets with more faces}|.
2. Increase the location loss weight and decrease the negative to
   positive ratio for |\textcolor{cyan}{datasets with more faces}|.
3. Use a lower learning rate and an optimizer |\textcolor{cyan}{such as}| Adam or SGD for
   datasets with |\textcolor{cyan}{facial landmarks}|.
\end{lstlisting}

\paragraph{Test task: FDDB-360}
\vphantom{X}

\begin{lstlisting}
1. For |\textcolor{cyan}{datasets with more faces}|, use a larger crop size and a higher
   anchor matching IoU threshold.
2. For |\textcolor{cyan}{datasets with fewer faces}|, use a smaller crop size and a lower
   anchor matching IoU threshold.
3. For datasets with no |\textcolor{cyan}{facial landmarks}|, use a lower location loss
   weight and a higher negative to positive ratio.
4. For datasets with |\textcolor{cyan}{facial landmarks}|, use a higher location loss
   weight and a lower negative to positive ratio.
5. For |\textcolor{cyan}{datasets with more faces}|, use a higher learning rate and an
   SGD optimizer.
6. For |\textcolor{cyan}{datasets with fewer faces}|, use a lower learning rate and an
   Adam optimizer.
\end{lstlisting}

\paragraph{Test task: MAFA}
\vphantom{X}

\begin{lstlisting}
1. Set the crop size and anchor matching IoU threshold according to
   the number of faces per image in the dataset: larger crop sizes and
   higher IoU thresholds for |\textcolor{cyan}{datasets with more faces}| per image, and
   smaller crop sizes and lower IoU thresholds for datasets with fewer
   faces per image.
2. Set the location loss weight according to the |\textcolor{cyan}{presence of facial}|
   |\textcolor{cyan}{landmarks in the dataset}|: higher weights for datasets with facial
   landmarks, and lower weights for datasets without |\textcolor{cyan}{facial landmarks}|.
3. Set the negative to positive ratio according to the |\textcolor{cyan}{difficulty of}|
   |\textcolor{cyan}{the dataset}|: higher ratios for datasets with more challenging
   scenarios (e.g. weather-based degradations, motion blur, focus blur).
4. Set the learning rate and optimizer according to the |\textcolor{cyan}{size of the}|
   |\textcolor{cyan}{dataset}|: lower learning rates and optimizers |\textcolor{cyan}{such as}| Adam or SGD for
   datasets with more images.
\end{lstlisting}

\paragraph{Test task: PASCAL VOC}
\vphantom{X}

\begin{lstlisting}
1. Set the crop size according to the |\textcolor{cyan}{number of faces in the dataset}|:
   larger crop sizes for |\textcolor{cyan}{datasets with more faces}|, and smaller crop
   sizes for |\textcolor{cyan}{datasets with fewer faces}|.
2. Set the anchor matching IoU threshold according to |\textcolor{cyan}{the number of}|
   |\textcolor{cyan}{faces}| in the dataset: higher thresholds for |\textcolor{cyan}{datasets with more faces}|,
   and lower thresholds for |\textcolor{cyan}{datasets with fewer faces}|.
3. Set the location loss weight according to the |\textcolor{cyan}{presence of facial}|
   |\textcolor{cyan}{landmarks in the dataset}|: lower weights for datasets with no facial
   landmarks, and higher weights for datasets with |\textcolor{cyan}{facial landmarks}|.
4. Set the negative to positive ratio according to |\textcolor{cyan}{the number of}|
   |\textcolor{cyan}{faces}| in the dataset: higher ratios for |\textcolor{cyan}{datasets with more faces}|, and
   lower ratios for |\textcolor{cyan}{datasets with fewer faces}|.
5. Set the learning rate and optimizer according to the |\textcolor{cyan}{difficulty of}|
   |\textcolor{cyan}{the dataset}|: higher learning rates and optimizers |\textcolor{cyan}{such as}| SGD for
   more challenging datasets.
\end{lstlisting}

\paragraph{Test task: UFDD}
\vphantom{X}

\begin{lstlisting}
1. Set the crop size and anchor matching IoU threshold according to
   the |\textcolor{cyan}{number of faces in the dataset}|: larger crop size and higher IoU
   threshold for |\textcolor{cyan}{datasets with more faces}|, smaller crop size and lower
   IoU threshold for |\textcolor{cyan}{datasets with fewer faces}|.
2. Set the location loss weight and negative to positive ratio
   according to the |\textcolor{cyan}{number of faces in the dataset}|: higher location loss
   weight and higher negative to positive ratio for |\textcolor{cyan}{datasets with more}|
   |\textcolor{cyan}{faces}|, lower location loss weight and lower negative to positive
   ratio for |\textcolor{cyan}{datasets with fewer faces}|.
3. Set the learning rate and optimizer according to the presence of
   |\textcolor{cyan}{facial landmarks}| in the dataset: lower learning rate and Adam
   optimizer for datasets with |\textcolor{cyan}{facial landmarks}|, higher learning rate
   and SGD optimizer for datasets without |\textcolor{cyan}{facial landmarks}|.
\end{lstlisting}

\paragraph{Test task: UMDAA-02}
\vphantom{X}

\begin{lstlisting}
1. Set the crop size according to the |\textcolor{cyan}{number of faces in the dataset}|:
   larger crop sizes for |\textcolor{cyan}{datasets with more faces}|, and smaller crop
   sizes for |\textcolor{cyan}{datasets with fewer faces}|.
2. Set the anchor matching IoU threshold according to |\textcolor{cyan}{the number of}|
   |\textcolor{cyan}{faces}| in the dataset: higher thresholds for |\textcolor{cyan}{datasets with more faces}|,
   and lower thresholds for |\textcolor{cyan}{datasets with fewer faces}|.
3. Set the location loss weight according to the |\textcolor{cyan}{presence of facial}|
   |\textcolor{cyan}{landmarks in the dataset}|: higher weights for datasets with facial
   landmarks, and lower weights for datasets without |\textcolor{cyan}{facial landmarks}|.
4. Set the negative to positive ratio according to |\textcolor{cyan}{the number of}|
   |\textcolor{cyan}{faces}| in the dataset: higher ratios for |\textcolor{cyan}{datasets with more faces}|, and
   lower ratios for |\textcolor{cyan}{datasets with fewer faces}|.
5. Set the learning rate and optimizer according to the |\textcolor{cyan}{difficulty of}|
   |\textcolor{cyan}{the dataset}|: higher learning rates and optimizers |\textcolor{cyan}{such as}| SGD or Adam
   for more challenging datasets.
\end{lstlisting}

\paragraph{Test task: WIDER FACE}
\vphantom{X}

\begin{lstlisting}
1. Set the crop size to a value that is proportional to |\textcolor{cyan}{the number of}|
   |\textcolor{cyan}{faces}| in the dataset.
2. Set the anchor matching IoU threshold to a value that is
   proportional to the |\textcolor{cyan}{number of faces in the dataset}|.
3. Set the negative to positive ratio to a value that is proportional
   to the |\textcolor{cyan}{number of faces in the dataset}|.
4. Set the learning rate to a value that is proportional to the
   |\textcolor{cyan}{number of faces in the dataset}|.
5. If the dataset contains |\textcolor{cyan}{facial landmarks}|, set the location loss
   weight to a value that is proportional to the number of faces in the
   dataset.
\end{lstlisting}

\paragraph{Test task: WIDER-FACE-360}
\vphantom{X}

\begin{lstlisting}
1. Set the crop size and anchor matching IoU threshold according to
   the |\textcolor{cyan}{number of faces in the dataset}|: larger crop size and higher IoU
   threshold for |\textcolor{cyan}{datasets with more faces}|, and smaller crop size and
   lower IoU threshold for |\textcolor{cyan}{datasets with fewer faces}|.
2. Set the location loss weight according to the presence of facial
   landmarks: higher weight for datasets with |\textcolor{cyan}{facial landmarks}|, and
   lower weight for datasets without |\textcolor{cyan}{facial landmarks}|.
3. Set the negative to positive ratio according to |\textcolor{cyan}{the number of}|
   |\textcolor{cyan}{faces}| in the dataset: higher ratio for |\textcolor{cyan}{datasets with more faces}|, and
   lower ratio for |\textcolor{cyan}{datasets with fewer faces}|.
4. Set the learning rate according to the number of faces in the
   dataset: higher rate for |\textcolor{cyan}{datasets with more faces}|, and lower rate for
   |\textcolor{cyan}{datasets with fewer faces}|.
5. Set the optimizer according to the |\textcolor{cyan}{number of faces in the dataset}|:
   SGD for |\textcolor{cyan}{datasets with more faces}|, and Adam for datasets with fewer
   faces.
\end{lstlisting}

\paragraph{Test task: WIKI}
\vphantom{X}

\begin{lstlisting}
1. Set the crop size to a value that is proportional to |\textcolor{cyan}{the number of}|
   |\textcolor{cyan}{faces}| in the dataset.
2. Set the anchor matching IoU threshold to a value that is
   proportional to the |\textcolor{cyan}{number of faces in the dataset}|.
3. Set the location loss weight to a value that is proportional to
   the presence of |\textcolor{cyan}{facial landmarks}| in the dataset.
4. Set the learning rate to a value that is inversely proportional to
   the |\textcolor{cyan}{negative to positive ratio in the dataset}|.
5. Use an optimizer |\textcolor{cyan}{such as}| Adam or SGD.
\end{lstlisting}

\end{document}